\pdfoutput=1

\documentclass[11pt]{article}

\usepackage{emnlp2021}

\usepackage{times}
\usepackage{latexsym}
\usepackage{graphicx}

\usepackage{microtype}
\usepackage{booktabs}
\usepackage{multirow}
\usepackage{amsmath}
\usepackage{amssymb}
\usepackage{array}
\usepackage{xspace}
\usepackage{bm}

\usepackage{color, colortbl}
\definecolor{Gray}{gray}{0.93}
\usepackage{array}
\newcolumntype{P}[1]{>{\raggedright\arraybackslash}p{#1}}
\newcolumntype{N}[1]{>{\raggedright\arraybackslash\columncolor{Gray}}p{#1}}

\usepackage[T1]{fontenc}

\usepackage[utf8]{inputenc}

\usepackage{microtype}

\newcommand\ftnote[1]{\footnote{\raggedright#1}}

\newcommand{\graphadapter}{{\small\textsc{StructAdapt}}\xspace}
\newcommand{\graphadaptergcn}{{\small\textsc{StructAdapt-gcn}}\xspace}
\newcommand{\graphadapterrgcn}{{\small\textsc{StructAdapt-rgcn}}\xspace}
\newcommand{\vanilladapter}{{\small\textsc{Adapt}}\xspace}
\newcommand{\finetune}{{\small\textsc{Fine-tune}}\xspace}
\newcommand{\finetunetop}{{\small\textsc{FT-top2}}\xspace}
\newcommand{\finetunebottom}{{\small\textsc{FT-bottom2}}\xspace}
\newcolumntype{P}[1]{>{\centering\arraybackslash}p{#1}}
\newcolumntype{M}[1]{>{\centering\arraybackslash}m{#1}}

\title{Structural Adapters in Pretrained Language Models \\ for AMR-to-Text Generation}

\author{Leonardo F. R. Ribeiro$^{\dag}$, Yue Zhang$^{\ddag}$ and Iryna Gurevych$^{\dag}$ \vspace{1mm} \\
\rule{0pt}{2.5ex}
  $^{\dag}$Ubiquitous Knowledge Processing Lab, Technical University of Darmstadt\\
  $^{\ddag}$School of Engineering, Westlake University \\
 \texttt{ribeiro@aiphes.tu-darmstadt.de}
}

\begin{document}
\maketitle
\begin{abstract}

Pretrained language models (PLM) have recently advanced graph-to-text generation, where the input graph is linearized into a sequence and fed into the PLM to obtain its representation. However, efficiently encoding the graph structure in PLMs is challenging because such models were pretrained on natural language, and modeling structured data may lead to catastrophic forgetting of distributional knowledge. In this paper, we propose \graphadapter, an adapter method to encode graph structure into PLMs. Contrary to prior work, \graphadapter effectively models interactions among the nodes based on the graph connectivity, only training graph structure-aware adapter parameters. In this way, we incorporate task-specific knowledge while maintaining the topological structure of the graph. We empirically show the benefits of explicitly encoding graph structure into PLMs using \graphadapter, outperforming the state of the art on two AMR-to-text datasets, training only 5.1\% of the PLM parameters.\ftnote{Our code and checkpoints are available at \href{https://github.com/UKPLab/StructAdapt}{https://github.com/UKPLab/StructAdapt}.}
\end{abstract}

\section{Introduction}

\begin{figure}[t]
    \centering
    \includegraphics[width=.47\textwidth]{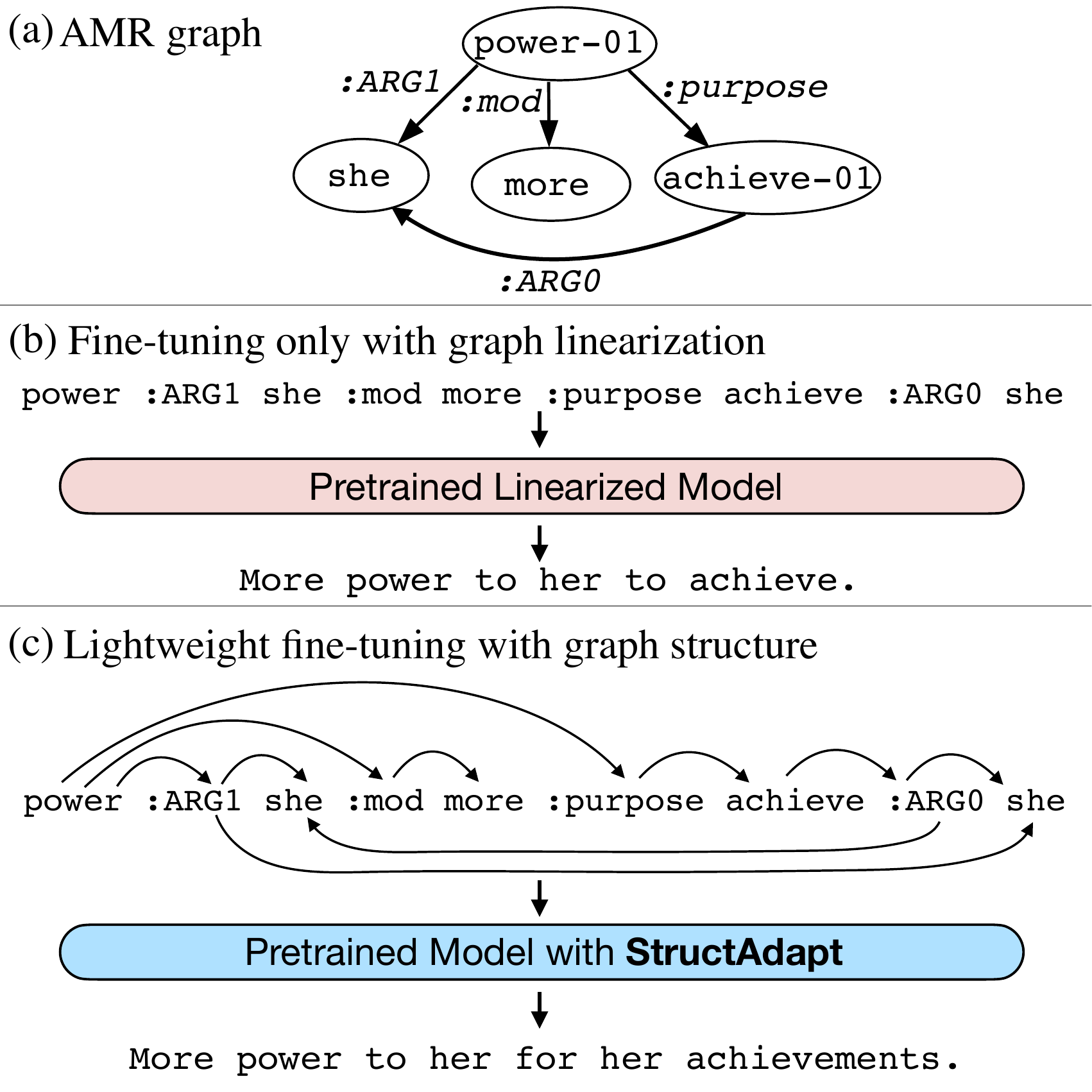}
    \caption{(a) AMR for the sentence \textit{More power to her for her achievements}. While in (b) the pretrained model gets as input the graph linearization, in (c) it additionally receives the graph connectivity information.}
    \label{fig:amrexamples}
\end{figure}

Data-to-text tasks aim to generate meaningful and coherent natural language text that faithfully conveys \emph{structured data}. Some examples of structured information include tables \cite{parikh-etal-2020-totto}, Knowledge Graphs (KGs) \cite{gardent-etal-2017-webnlg, VOUGIOUKLIS20181} and Abstract Meaning Representation (AMR) \cite{banarescu-etal-2013-abstract}. In this work, we focus on AMR-to-text generation where the goal is to generate a fluent and grammatical sentence that is faithful to a given AMR graph (See Figure~\ref{fig:amrexamples}a). AMR is a semantic formalism that has received much research interest \cite{song-etal-2018-graph, doi:10.116200269, ribeiro-etal-2019-enhancing, opitz-etal-2020-amr, bamboo, fu-etal-2021-end} and has been shown to benefit downstream tasks such as text summarization \cite{liao-etal-2018-abstract} and machine translation \cite{doi:10.116200252}.  Both statistical \cite{flanigan-etal-2016-generation, pourdamghani-etal-2016-generating} and neural methods \cite{bai-etal-2020-online,cai-lam-2020-graph} have been investigated for AMR-to-text generation, and dominant methods make use of Graph Neural Networks (GNNs) \cite{Kipf:2016tc} or Transformers \cite{NIPS2017_7181} for representing the input graph.

\emph{Pretrained language models} (PLMs) \cite{devlin-etal-2019-bert, liu2020roberta, radford2019language, lewis2019bart} have been shown useful as a general text representation method, giving much improved results on a wide range of tasks \cite{wang-etal-2018-glue, NIPS2019_8589}. However, they cannot be directly leveraged to benefit AMR-to-text generation, and more generally graph-to-text generation, due to the structural nature of the input. One solution is to transform the structured input into a sequence, which can be directly fed into PLMs (See Figure~\ref{fig:amrexamples}b). Recent studies \cite{mager2020gpttoo, harkous2020text,ribeiro2020investigating,ribeiro2021smelting} transform AMRs into sequences by top-down linearization \cite{konstas-etal-2017-neural}. It has been shown that such linearized graph representation can be used to fine-tune a PLM and improve graph-to-text generation performances \cite{kale2020texttotext}.

The above methods, however, suffer from two salient limitations. First, linearized graph structures are different in nature from natural language. As a result, knowledge from large-scale pretraining intuitively cannot be fully transferred, and fine-tuning a sentence representation using linearized graphs can lead to catastrophic forgetting of such distributional knowledge \cite{goodfellow2013an, Kirkpatrick3521}. Second, a linearized representation weakens structural information in the original graphs by diluting the explicit connectivity information (i.e., which nodes are connected to each other), and PLMs must infer how edge connections are specified in the sequence. This fact was also observed by \citet{song-etal-2018-graph}, \citet{beck-etal-2018-graph} and \citet{ribeiro-etal-2019-enhancing}, who show that GNN encoders outperform sequential encoders for AMR-to-text generation without pretraining.

To mitigate the issues, we aim to explicitly encode the graph data into a PLM without contaminating its original distributional knowledge. To this end, we propose \graphadapter, a novel structure-aware adapter that effectively allows leveraging the input graph structure into PLMs (See Figure~\ref{fig:amrexamples}c). The main idea is to add layer-wise modules, which extract information from the pretrained layers and make use of it in a graph-structure encoding. As shown in Figure~\ref{fig:adapterarc}, \graphadapter employs a \emph{graph convolution} in order to learn representations built upon the graph connectivity over the PLM encoder. Because \graphadapter is added to each encoder layer, deep integration of linguistic knowledge and graph knowledge can be achieved. During fine-tuning, only the adapter parameters are trained, whereas the PLM parameters remain unchanged, in contrast to previous methods based on the graph linearizations that fine-tune all model parameters.

Empirically we show that \graphadapter significantly outperforms linearized fine-tuning baselines and naive sequential adapters \cite{pmlr-v97-houlsby19a}. Moreover, \graphadapter is more robust to different graph linearizations, better treats reentrancies (nodes with more than one entering edge) and long-range node dependencies. Our proposed models, based on \graphadapter, surpass the current state of the art on LDC2017T10 and LDC2020T02 datasets by up to 3.1 BLEU points, training only 5.1\% of the original PLM parameters.

\section{Related Work}

\paragraph{Fine-tuning for Graph-to-text Generation.} While previous approaches \cite{song-etal-2018-graph,ribeiro-etal-2019-enhancing, cai-lam-2020-graph, schmitt2020modeling, zhang-etal-2020-lightweight} have shown that explicitly encoding the graph structure is beneficial, fine-tuning PLMs on linearized structured data has established a new level of performance in data-to-text generation \cite{radev2020dart, kale2020texttotext, ribeiro2021smelting}. Our work can be seen as integrating the advantage of both graph structure encoding and PLMs, using a novel adapter module. 

\citet{mager2020gpttoo} employ cycle consistency to improve the adequacy of generated texts from AMRs using GPT-2 \cite{radford2019language}, whereas \citet{harkous2020text} train a classifier to rank candidate generations based on the semantic fidelity. \citet{ribeiro2020investigating} investigate encoder-decoder PLMs for graph-to-text generation, and show that task-specific pretraining can lead to notable improvements and that PLMs benefit much more from the graph structure of AMRs than of KGs. \citet{hoyle2020promoting} explore the extent to which PLMs are invariant to graph linearization, finding that models trained on canonical linearizations fail to generalize to meaning-preserving alternatives. Compared to this line of work, which tunes all PLM parameters, our method obtains a further 19x reduction in task-specific parameters, tuning only 5.1\% of the parameters while achieving state-of-the-art performance, being more robust to permutations of the graph representation and better encoding larger graphs.

\paragraph{Lightweight Fine-tuning.}
Recently, different approaches have emerged as an alternative training strategy in order to avoid fine-tuning all parameters of a PLM. \citet{sidetuning2019} train a lightweight ``side'' network that is fused with the pretrained model via summation. \citet{li2021prefixtuning} propose to prepend a trainable continuous prefix as an alternative to adapters, maintaining comparable performance in data-to-text tasks using fewer trained parameters. \citet{DBLP:journals/corr/abs-2103-10385} develop a method to automatically search prompts in the continuous space and evaluate it in few-shot NLU tasks. \citet{hambardzumyan-etal-2021-warp} propose adversarial reprogramming attempts to learn task-specific word embeddings to customize the language model for the downstream task.

Adapter-based approaches \cite{pmlr-v97-houlsby19a, NIPS2017_e7b24b11,lauscher-etal-2020-common,pfeiffer-etal-2020-adapterhub, pfeiffer2021adapterfusion} introduce a small number of task specific parameters, keeping the underlying pretrained model fixed. \citet{pfeiffer-etal-2020-mad} propose an adapter method to arbitrary tasks and languages by learning modular language and task representations. The above works are related to \graphadapter as it trains much fewer parameters, but also different because they do not explicitly encode the input structure, whereas \graphadapter directly aims to encode it.

\section{Graph-to-Text Model}

Let $\mathcal{G}_0 = (\mathcal{V}_0, \mathcal{E}_0, \mathcal{R}_0)$ denote a rooted and directed AMR graph with a node set $\mathcal{V}_0$ and labeled edges $ (u, r, v) \in \mathcal{E}_0$, where $u, v \in \mathcal{V}_0$ and $r \in \mathcal{R}_0$ is a relation type. An example of an AMR graph and its corresponding sentence is shown in Figure~\ref{fig:amrexamples}a.

\subsection{Encoder-Decoder Architecture}
\label{sec:baselinemodel}
Consider a conditional generation task where the input is a context $x$ and the output $y = \langle y_1,\dots,y_{|y|} \rangle$ is a sequence of tokens. In AMR-to-text generation, the context $x$ is the AMR graph and $y$ is the sentence that describes the AMR graph in natural language. 

Let $p_\phi(y \mid x)$ denote a PLM parametrized by $\phi$, where $x$ is encoded by a bidirectional encoder, and the decoder predicts $y$ autoregressively, conditioned on the encoded $x$ and its left context. We focus on PLMs based on the Transformer encoder-decoder architecture \cite{NIPS2017_7181}, as they are suitable for conditional text generation. We define $x = \textsc{LIN}(\mathcal{G}_0)$, where $\textsc{LIN}$ is a function that linearizes $\mathcal{G}_0$ into a sequence of tokens.\footnote{The variable of a re-entrant node -- node with more than one incoming edge -- is replaced with its co-referring concept.} Following~\citet{damonte-cohen-2019-structural}, as shown in Figure~\ref{fig:amrexamples}b, we linearize the AMR into a sequence of nodes and edges using the depth-first traversal of the canonical human-created AMR.\footnote{Other AMR linearizations are discussed in \S\ref{sec:amrinputrep}.} In a nutshell, the hidden representation $\mathbf{h}^{l}_i \in \mathbb{R}^{d}$, for all $x_i \in x$, is computed by the encoder layer $l$, where $d$ is the hidden dimension. The decoder hidden representation $\hat{\mathbf{h}}_{i}^{l} \in \mathbb{R}^{d}$ is computed by the layer $l$ of the autoregressive decoder at time step $i$.

\subsection{Fine-tuning}
The model is initialized with pretrained parameters $\phi$ (e.g. using T5, \citeauthor{2019t5}, \citeyear{2019t5}) and fine-tuned to optimize the following log-likelihood objective over each gold instance $(x, y)$:
\begin{align}
    \max_{\phi} ~\log p_\phi(y \mid x) = \sum_{i=1}^{|y|}  \log p_\phi(y_i \mid y_{1:i-1}, x) \text{.}
    \label{eqn:loss}
\end{align}

\begin{figure}[t]
    \centering
    \includegraphics[width=.4\textwidth]{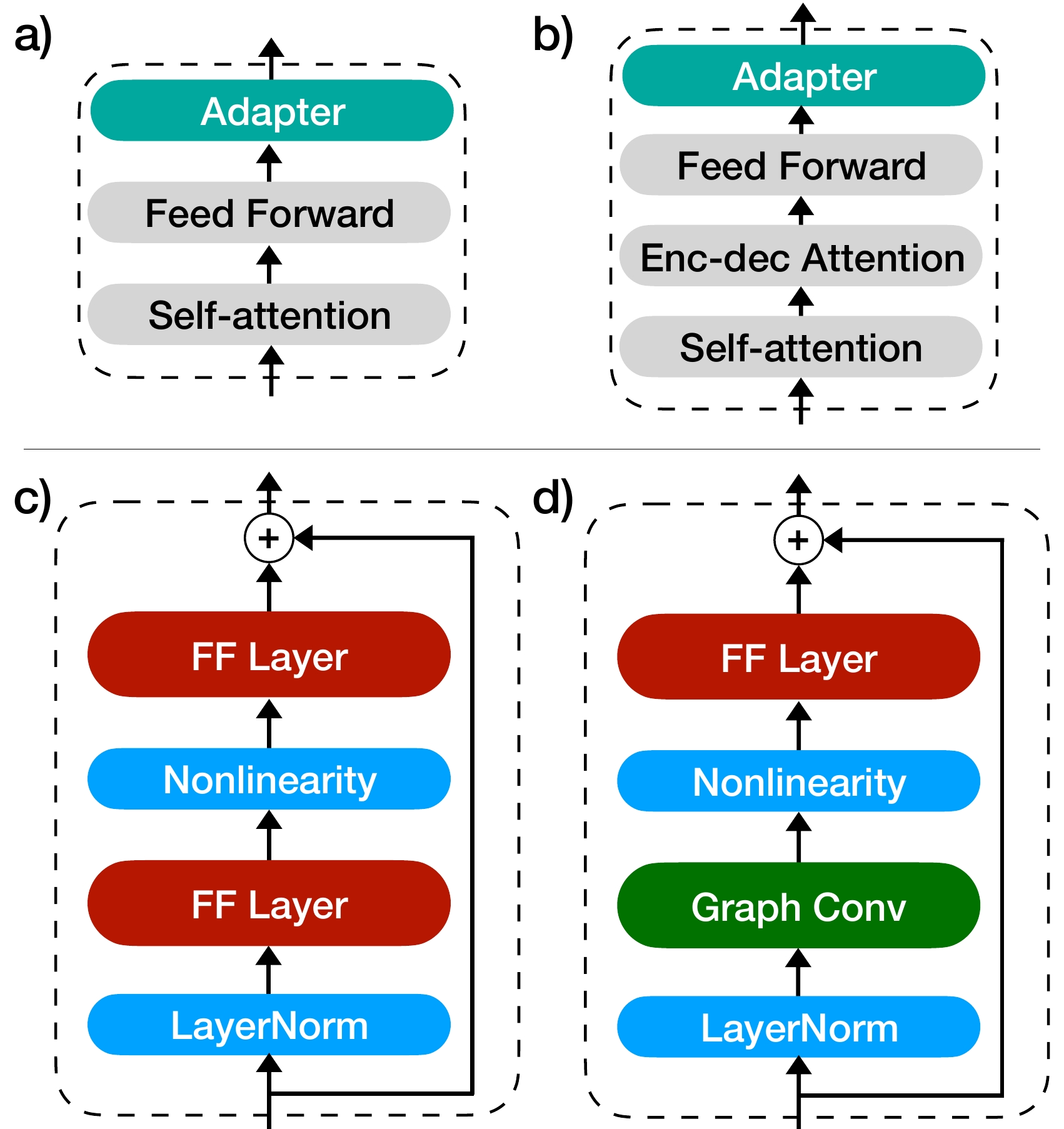}
    \caption{Integration of the adapter modules with the (a) encoder and (b) decoder layers of the Transformer; layer normalization and residual connections are omitted for clarification. (c) \vanilladapter with two feed-forwards layers. (d) \graphadapter encodes the graph structure using a \emph{graph convolutional layer}.}
    \label{fig:adapterarc}
\end{figure}

\subsection{Baseline Adapter} We employ an adapter module after the feed-forward sub-layer of each layer on both encoder (Figure~\ref{fig:adapterarc}a) and decoder (Figure~\ref{fig:adapterarc}b) of the PLM. We modify the adapter architecture from \citet{pmlr-v97-houlsby19a}, computing the adapter representation at each layer $l$, given the encoder layer representation $\mathbf{h}_{i}^{l}$ (or $\hat{\mathbf{h}}_{i}^{l}$ in the decoder), as follows:
\begin{align}
\hat{\mathbf{z}}_i &= \mathbf{W}_o^{l} ( \sigma (\mathbf{W}^{l}_p \,  {\scriptstyle\mathsf{LN}}(\mathbf{h}^{l}_i))) + \mathbf{h}^{l}_i \, ,
\end{align}
where $\sigma$ is the activation function and ${\scriptstyle\mathsf{LN}}(\cdot)$ denotes layer normalization. $\mathbf{W}^{l}_o \in \mathbb{R}^{d \times m}$ and $\mathbf{W}^{l}_p \in \mathbb{R}^{m \times d}$ are adapter parameters, and $m$ is the hidden dimension of the adapter.
Figure~\ref{fig:adapterarc}c illustrates the baseline adapter module, which we call \vanilladapter. 

\paragraph{Training.} Let the set of adapters' parameters for the encoder and decoder layers be parametrized by $\theta$. The training objective is the same as Equation (\ref{eqn:loss}), but the set of trainable parameters changes: the PLM parameters $\phi$ are frozen and the adapter parameters $\theta$ are the only trainable parameters. In contrast to fine-tuning, adapters substantially reduce the number of trainable parameters that are used to adapt the PLM to the downstream task.

\subsection{Limitation}
\label{sec:intuition}

Intuitively, the connection between nodes in the input graph can influence the encoding of $x$ by guiding what to extract from $x$ in order to generate $y$. Note that in both fine-tuning and \vanilladapter approaches, the self-attention mechanisms of the encoder layers treat the sequence of nodes and edges $x$ essentially as a fully connected graph, greatly diluting the original graph structure. In this way, the model has to retrieve the original connectivity of the graph from $x$. For example, the AMR linearization in Figure~\ref{fig:amrexamples}b has two mentions of the node \emph{she}, and the model should capture that both mentions belong to the same node in the original graph. 

\section{Structural Adapter}

We propose \graphadapter, a lightweight alternative to injecting \emph{structural inductive bias}\footnote{The model architecture explicitly encodes the graph structure, i.e., which nodes are connected to each other.} into PLMs.

We first describe the intuition in \S\ref{sec:intuition} and define our method formally in \S\ref{sec:method}.

\subsection{Intuition}
\label{sec:intuition}

Injecting \emph{graph structural bias} into graph-to-text models trained from scratch improves the performance compared to linearized approaches \cite{damonte-cohen-2019-structural, ribeiro-etal-2019-enhancing}. However, it is not straightforward how to effectively model the input graph structure when fine-tuning PLMs, which usually are pretrained using natural language and not structured data. 

Our key idea is modeling the graph connectivity in the encoder utilizing an adapter module, using information flows between adjacent nodes in a message-passing update, employing a \emph{graph convolution} (see Figure~\ref{fig:adapterarc}d). In this way, the graph structure substantially impacts the node representations, better encoding the input graph without impacting the knowledge learned during pretraining. This can lead to more efficient and better AMR-to-text generation as we will show in \S\ref{sec:exps} and \S\ref{sec:graphrepeval}. Moreover, different adapters for distinct graph domains can be used with the same PLM, yielding a high degree of parameter sharing for graph-to-text tasks.

\subsection{Graph Representation}
\label{sec:graphrep}
\begin{figure}[t]
    \centering
    \includegraphics[width=.45\textwidth]{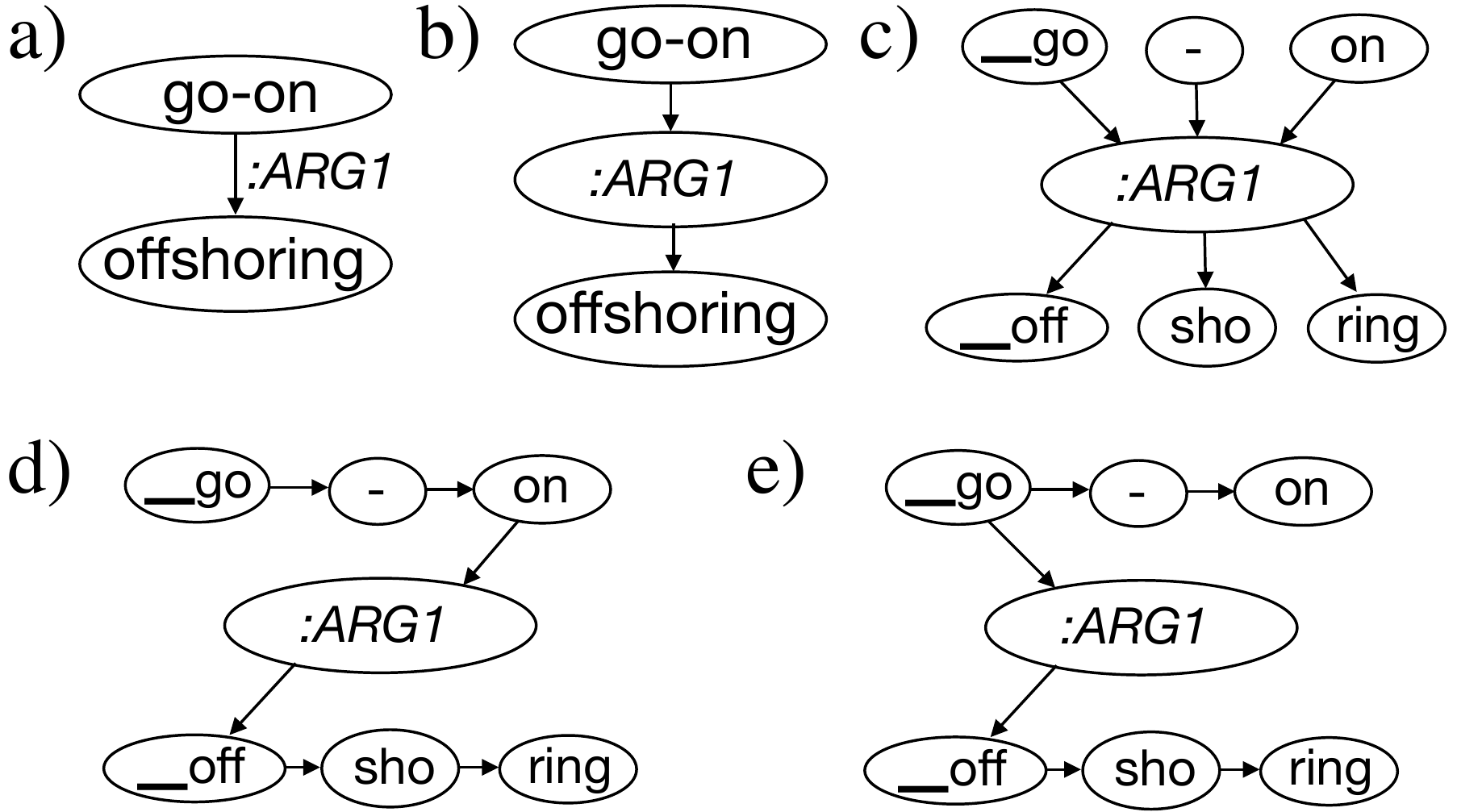}
    \caption{An example of (a) an AMR graph structure, (b) its unlabeled version and three different subword representations: (c) \emph{rep1}, (d) \emph{rep2} and (e) \emph{rep3}.}
    \label{fig:amrreps}
\end{figure}

We convert each $\mathcal{G}_0$ into a bipartite graph $\mathcal{G}_1 = (\mathcal{V}_1, \mathcal{E}_1)$, replacing each labeled edge $(u,r,v) \in \mathcal{E}_0$ with two unlabeled edges $e_1 = (u, r)$ and $e_2 = (r, v)$. Similar to~\citet{beck-etal-2018-graph}, this process converts the graph into its unlabeled version. Figure~\ref{fig:amrreps} shows an (a) AMR subgraph and (b) its unlabeled representation.

Note that PLMs typically use a vocabulary with subword units \cite{sennrich-etal-2016-neural}. This presents a challenge in how to represent such a graph using subword tokens. Inspired by~\citet{ribeiro2020modeling}, we transform each $\mathcal{G}_1$ into a new token graph $\mathcal{G} = (\mathcal{V}, \mathcal{E})$, where each token of a node in $\mathcal{V}_1$ becomes a node $v \in \mathcal{V}$. We convert each edge $ (u_1, v_1) \in \mathcal{E}_1$ into a set of edges and connect every token of $u_1$ to every token of $v_1$. That is, an edge $(u, v)$ will belong to $\mathcal{E}$ if and only if there exists an edge $ (u_1, v_1) \in \mathcal{E}_1$ such that $u \in u_1$ and $v \in v_1$, where $u_1$ and $v_1$ are seen as sets of tokens. Figure~\ref{fig:amrreps}c shows an example of the token graph.

\subsection{Method}
\label{sec:method}

\graphadapter employs a two-layer architecture in order to re-purpose the PLM for the graph-to-text task using a small number of new parameters. Formally, for each node $v \in \mathcal{V}$, given the hidden representation $\mathbf{h}^{l}_v$ from the encoder layer $l$, \graphadapter computes:
\begin{align}
\mathbf{g}^{l}_{v} &= {\mathsf{GraphConv}}_{l} ({\scriptstyle\mathsf{LN}}(\mathbf{h}^{l}_v),\!\{ {\scriptstyle\mathsf{LN}}(\mathbf{h}^{l}_u)\!: u \in {\scriptstyle\mathcal{N}(v)} \})\hspace{-.5em}  \nonumber \\
\mathbf{z}^{l}_v &= \mathbf{W}^{l}_{e} \sigma (\mathbf{g}^{l}_{v}) + \mathbf{h}^{l}_v \, ,
\end{align}
where $\mathcal{N}(v)$ is the immediate neighborhood of $v$ in $\mathcal{G}$. $\mathsf{GraphConv}_{l}(\cdot)$ is the graph convolution that computes the node representation based on the \emph{local neighborhood} of $v$, and $\mathbf{W}^{l}_e \in \mathbb{R}^{d \times m}$ is a parameter. Figure~\ref{fig:adapterarc}d illustrates \graphadapter.\footnote{Preliminary experiments with other architecture configurations led to worse or similar performance.}

\paragraph{Graph Convolution.} The graph convolutional layer allows exploration of distinct strategies for neighborhood aggregation in order to model structural information of the input graph. Different GNN architectures \cite{velickovic2018graph, xu2018how} can be employed as the graph convolution. Moreover, in this way, we avoid changing the self-attention mechanism of the current pretrained encoder, allowing to also capture \emph{global information} based on the pretrained knowledge.

Our graph convolution is based on the Graph Convolutional Network (GCN) proposed by~\citet{Kipf:2016tc}. At each layer $l$, we compute the representation of a node $v \in \mathcal{V}$ as follows:  

\begin{equation}
        \mathbf{g}^{l}_{v} =
        \sum_{u \in \mathcal{N}(v)} \frac{1}{\sqrt{d_{v}d_{u}}}  \, \mathbf{W}^{l}_g \mathbf{h}^{l}_u  \, \,\,\text{,}
\label{eq:3}
\end{equation}
where $\mathcal{N}(v)$ is a set of nodes with incoming edges to $v$ and $v$ itself, $d_{v}$ is the degree of $v$, and $\mathbf{W}^{l}_g \in \mathbb{R}^{m \times d}$ is a parameter. 

We also consider the variant relational GCN (RGCN)~\cite{Schlichtkrull2018ModelingRD} as graph convolution. RGCN allows capturing the reverse edge direction so that we can consider the differences in the incoming and outgoing relations, which has shown to be beneficial~\cite{beck-etal-2018-graph}. In particular, the node representation is computed as:
\begin{equation}
        \mathbf{g}^{l}_{v} =
        \sum_{r \in \mathcal{R}} 
        \sum_{u \in \mathcal{N}_r(v)}   
        \frac{1}{|\mathcal{N}_r(v)|}
        \mathbf{W}^{l}_r \mathbf{h}^{l}_u  \, \,\,\text{,}
\label{eq:3}
\end{equation}
where $\mathcal{R}$ denotes the set of relations, i.e., the edge types \emph{default} and \emph{reverse}, $\mathcal{N}_r(v)$ denotes the set of neighbors under relation $r \in \mathcal{R}$, and $\mathbf{W}^{l}_{r} \in \mathbb{R}^{m \times d}$ encodes the edge type between the nodes $u$ and $v$.

Note that \graphadapter computes the refined structural node representation $\mathbf{z}^{l}_v$ based on the local node context, using as input the global representation $\mathbf{h}^{l}_v$ generated by the current PLM encoder layer. In this way, the model is able to capture both the global context based on the PLM linguistic knowledge and the local context based on the graph knowledge. Finally, we employ \vanilladapter into the decoder in order to adapt the language model to the graph-to-text task.

\section{Experiments}
\label{sec:exps}

Our models are initialized with pre-trained T5 \hbox{\cite{2019t5}}, but our approach can be combined with other PLMs such as BART \cite{lewis2019bart}. Our implementation is based on Hugging Face Transformer models \citep{wolf2019huggingfaces}. We use T5\textsubscript{base} for all experiments and report results with T5\textsubscript{large} for the test sets.\footnote{Hyperparameter details are in the appendix~\ref{appe:hyperparameters}.} We use the Adam optimizer \cite{kingma:adam} and employ a linearly decreasing learning rate schedule without warm-up. BLEU is used for the stopping criterion. Following recent work~\cite{mager2020gpttoo,zhang-etal-2020-lightweight}, we evaluate our proposed models on LDC2017T10 and LDC2020T02 corpora. 

\paragraph{Evaluation.} We evaluate the results with BLEU \cite{papineni-etal-2002-bleu} and chrF++ \cite{popovic-2015-chrf} metrics. We also report the meaning ($\mathcal{M}$) component of the $\mathcal{MF}$-score \cite{opitz-frank-2021-towards}, which measures how well the source AMR graph can be reconstructed from the generated sentence. We use BERTScore~\cite{bert-score} allowing a semantic evaluation that depends less on the surface forms. Finally, we also perform a human evaluation (\S\ref{sec:humanEval}).

\begin{table}[t]
\centering
\resizebox{\columnwidth}{!}{
{\renewcommand{\arraystretch}{0.9}

\begin{tabular}{@{\hspace*{1mm}}l@{\hspace*{3mm}}l@{\hspace*{3mm}}l@{\hspace*{3mm}}l@{\hspace*{3mm}}l@{\hspace*{1mm}}}
\toprule
 &\textbf{BLEU} &\textbf{chrF++}  & \,\,\, $\boldsymbol{\mathcal{M}}$ & \textbf{BERT}  \\
 \midrule
 
\citet{mager2020gpttoo} & 33.0 & 63.9 & \,\,\,\,\,\,\,- & \,\,\,\,\,\,\,-\\
\citet{zhang-etal-2020-lightweight} & 33.6 & 63.2 & \,\,\,\,\,\,\,- & \,\,\,\,\,\,\,-\\
\citet{harkous2020text} & 37.7 & \,\,\,\,\,- & \,\,\,\,\,\,\,- & \,\,\,\,\,\,\,-\\
\citet{hoyle2020promoting} & 44.9 & \,\,\,\,\,- & 76.54 & \,\,\,\,\,\,\,-\\
\citet{ribeiro2020investigating} & 45.8 & 72.5 & \,\,\,\,\,\,\,- & \,\,\,\,\,\,\,-\\[0.15cm]
\toprule
 \multicolumn{5}{c}{T5\textsubscript{base}}   \\
 \midrule
\finetune & 38.3{\small $\pm$0.3} & 68.6{\small $\pm$0.1} & 77.8{\small $\pm$0.3} & 95.5{\small $\pm$0.1}\\
\finetunetop{\small(14.8\%)} & 29.9{\small $\pm$0.1} & 63.0{\small $\pm$0.1} & 74.1{\small $\pm$0.2} & 94.4{\small $\pm$0.2}\\
\finetunebottom{\small(14.8\%)} & 35.9{\small $\pm$0.3} & 67.0{\small $\pm$0.2} & 76.9{\small $\pm$0.1} & 95.3{\small $\pm$0.1}\\
\vanilladapter{\small(8.5\%)} & 38.7{\small $\pm$0.4} & 69.2{\small $\pm$0.2} & 78.3{\small $\pm$0.1} & 95.6{\small $\pm$0.1}\\
\graphadapter{\small-GCN(2.1\%)} & 39.0{\small $\pm$0.3} & 69.1{\small $\pm$0.2} & 78.4{\small $\pm$0.2} & 95.7{\small $\pm$0.2}\\
\graphadapter{\small-GCN(8.5\%)} & 41.0{\small $\pm$0.5} & 70.0{\small $\pm$0.2} & 78.4{\small $\pm$0.1} & 95.7{\small $\pm$0.1}\\
\graphadapter{\small-RGCN(6.3\%)} & \textbf{44.0}{\small $\pm$0.3} &
\textbf{71.2}{\small $\pm$0.2} & \textbf{79.4}{\small $\pm$0.1} &
\textbf{95.9}{\small $\pm$0.2}\\

\midrule
\multicolumn{5}{c}{T5\textsubscript{large}}   \\

\midrule
\finetune & 41.2{\small $\pm$0.5} & 70.2{\small $\pm$0.2}& 78.0{\small $\pm$0.1} &95.8{\small $\pm$0.2}\\
\finetunetop{\small(7.9\%)} & 28.8{\small $\pm$0.4} & 61.8{\small $\pm$0.5} & 73.9{\small $\pm$0.2} & 94.1{\small $\pm$0.2}\\
\finetunebottom{\small(7.9\%)} & 37.6{\small $\pm$0.3} & 68.0{\small $\pm$0.2} & 77.2{\small $\pm$0.2} & 95.5{\small $\pm$0.1}\\
\vanilladapter{\small(6.8\%)} & 42.9{\small $\pm$0.3} & 71.6{\small $\pm$0.2} & 78.9{\small $\pm$0.1} & 96.1{\small $\pm$0.1}\\
\graphadapter{\small-GCN(1.7\%)} & 44.1{\small $\pm$0.4} & 71.8{\small $\pm$0.3} & 79.1{\small $\pm$0.1} & 96.1{\small $\pm$0.2}\\
\graphadapter{\small-GCN(6.8\%)} & 45.8{\small $\pm$0.2} & 72.5{\small $\pm$0.1} & 79.3{\small $\pm$0.2} & 96.2{\small $\pm$0.1}\\
\graphadapter{\small-RGCN(5.1\%)} & \textbf{46.6}{\small $\pm$0.3} & \textbf{72.9}{\small $\pm$0.2} &\textbf{79.6}{\small $\pm$0.1} & \textbf{96.3}{\small $\pm$0.1}\\
\bottomrule
\end{tabular}}}
\caption{Results on the LDC2017T10 test set. Mean ($\pm$s.d.) over 4 seeds.}
\label{tab:testsetresults}
\end{table}

\subsection{Main Results}

We compare \graphadapter with four methods: fine-tuning (\finetune), fine-tuning only the top or bottom 2 layers (\finetunetop, \finetunebottom) and \vanilladapter. All models use the same graph linearization generated by the depth-first traversal. We also report recent state-of-the-art results on both datasets. Tables~\ref{tab:testsetresults} and \ref{tab:testsetresults-ldc2020} show the results.

We find that training only 5.1\% task-specific parameters, \graphadapterrgcn achieves a BLEU score of 46.6 in LDC2017T10, substantially improving over \finetune and other lightweight baselines (\vanilladapter, \finetunetop, \finetunebottom), and outperforming \citet{ribeiro2020investigating} and \citet{hoyle2020promoting} which fine-tune T5 updating significantly more parameters. \graphadapter also achieves state-of-the-art performance on LDC2020T02, considerably improving over \citet{Micheleamr}, which implicitly models the graph structure information using linearization techniques.

In general, \graphadapter is better than \vanilladapter when training the same number of parameters, and slightly better even when training only 1.7\% of the parameters for both datasets. This highlights that the gains not only come from using an adapter architecture, but from considering the graph connectivity. \graphadapterrgcn is more effective than \graphadaptergcn using fewer parameters, demonstrating that considering reverse relations is advantageous. \vanilladapter is consistently better than \finetune, agreeing with our intuition of catastrophic forgetting when fine-tuning. Interestingly, in contrast to popular strategies that focus on upper layers in fine-tuning~\cite{howard-ruder-2018-universal, pmlr-v97-houlsby19a, li2021prefixtuning}, \finetunebottom's performance is better than \finetunetop's, suggesting that lower layers have a significant impact in adapting the PLM to structured data.

\begin{table}[t]
\centering
\resizebox{\columnwidth}{!}{
{\renewcommand{\arraystretch}{0.9}

\begin{tabular}{@{\hspace*{1mm}}l@{\hspace*{3mm}}l@{\hspace*{3mm}}l@{\hspace*{3mm}}l@{\hspace*{3mm}}l@{\hspace*{1mm}}}
\toprule
 &\textbf{BLEU} &\textbf{chrF++}  & \,\,\, $\boldsymbol{\mathcal{M}}$ &\textbf{BERT}  \\
 \midrule

\citet{zhang-etal-2020-lightweight} & 34.3 & 63.7 & \,\,\,\,\,\,\,- & \,\,\,\,\,\,\,-\\
\citet{Micheleamr} & 44.9 & 72.9 & \,\,\,\,\,\,\,- & \,\,\,\,\,\,\,-\\

\midrule
\multicolumn{4}{c}{T5\textsubscript{large}}   \\
\midrule
\finetune & 41.6{\small $\pm$0.6} & 70.4{\small $\pm$0.5} & 78.5{\small $\pm$0.2} & 96.0{\small $\pm$0.1}\\
\finetunetop{\small(7.9\%)} & 33.4{\small $\pm$0.5} &  63.5{\small $\pm$0.3} & 73.4{\small $\pm$0.4}& 94.3{\small $\pm$0.1}\\
\finetunebottom{\small(7.9\%)} & 38.2{\small $\pm$0.2} &  68.3{\small $\pm$0.1} & 78.1{\small $\pm$0.2}& 95.6{\small $\pm$0.1}\\
\vanilladapter{\small(6.8\%)} & 43.0{\small $\pm$0.2} & 71.3{\small $\pm$0.2} & 79.3{\small $\pm$0.1} & 96.2{\small $\pm$0.1}\\
\graphadapter{\small-GCN(1.7\%)} & 46.2{\small $\pm$0.2} & 71.8{\small $\pm$0.2} & 79.4{\small $\pm$0.3}  & 96.0{\small $\pm$0.2}\\
\graphadapter{\small-GCN(6.8\%)}& 47.1{\small $\pm$0.4} & 72.5{\small $\pm$0.1}  & 79.7{\small $\pm$0.2} & 96.2{\small $\pm$0.1} \\
\graphadapter{\small-RGCN(5.1\%)} & \textbf{48.0}{\small $\pm$0.2} & \textbf{73.2}{\small $\pm$0.1} & \textbf{80.1}{\small $\pm$0.3} & \textbf{96.3}{\small $\pm$0.1}\\
\bottomrule
\end{tabular}}}
\caption{Results on the LDC2020T02 test set.}
\label{tab:testsetresults-ldc2020}
\end{table}

Different from our work, both \citet{mager2020gpttoo} and \citet{ribeiro2020investigating} use the {\small\textsc{penman}} notation which makes the input much longer (containing more tokens), and demonstrate that this representation is able to achieve strong results -- this is orthogonal to our \graphadapter representation and can be incorporated in future work. 

\begin{table}[t]
\small
\centering
{\renewcommand{\arraystretch}{0.9}
\setlength\tabcolsep{2pt}
\setlength{\belowrulesep}{1pt}
\setlength{\aboverulesep}{1pt}
\begin{tabular}{c c c c} 
\toprule
\textbf{Graph Size} & \multicolumn{1}{c}{\vanilladapter} & \multicolumn{1}{c}{\graphadapterrgcn} &  \\
\midrule
 {\small All} & ${5.6}^{{\scriptscriptstyle A}}$  & ${6.1}^{{\scriptscriptstyle B}}$ &\\
\midrule
01-30 & ${6.1}^{{\scriptscriptstyle A}}$ & ${6.2}^{{\scriptscriptstyle A}}$ &\\
31-60 & ${5.4}^{{\scriptscriptstyle A}}$ & ${5.4}^{{\scriptscriptstyle A}}$ &\\
${>}60$ & ${5.2}^{{\scriptscriptstyle A}}$ & ${6.2}^{{\scriptscriptstyle B}}$ &\\
\bottomrule
\end{tabular}}

\caption{Meaning similarity obtained in the human evaluation. The ranking was determined by Mann-Whitney tests with $p{<}0.05$. Difference between systems which have a letter in common is not statistically significant.}
\label{tab:humanevevaluation}
\end{table}

Overall, the results indicate that explicitly considering the graph structure using an adapter mechanism is effective for AMR-to-text generation, significantly reducing the number of trained parameters while improving generation quality.

\subsection{Human Evaluation}
\label{sec:humanEval}
To further assess the quality of the generated texts by the adapter-based models in LDC2020T02, we conduct a human evaluation via crowdsourcing using Amazon Mechanical Turk.  We follow previous work~\cite{ribeiro-etal-2019-enhancing,castro-ferreira-etal-2019-neural} and evaluate the \emph{meaning similarity}, i.e., how close in meaning is the generated text to the reference sentence.\footnote{We also assessed the \emph{fluency} of the texts and the differences between the models were not statistically significant.} We divide the datapoints into 3 different sets by by the graph size, i.e., the number of nodes, after converting edges into nodes (cf.\ \S\ref{sec:graphrep}). This setting allows us to evaluate the performance of the models based on the complexity of the AMR graph.

We randomly select 100 generated texts for each set and each model (total of 600), which annotators then rate on a 1-7 Likert scale. For each text we collect scores from 3 annotators and use MACE~\cite{hovy-etal-2013-learning}, a Bayesian model that incorporates the reliability of individual workers, to merge sentence-level labels.\footnote{Refer to Appendix~\ref{appe:humaneval} for a detailed description of the human evaluation.} Table~\ref{tab:humanevevaluation} shows that \graphadapter improves the meaning similarity over \vanilladapter with statistically significant margins ($p{<}0.05$). Note that the gains mainly come from datapoints with ${>}60$ nodes, indicating that \graphadapter is better when encoding larger graphs.

\begin{figure}[t]
    \includegraphics[width=0.48\textwidth]{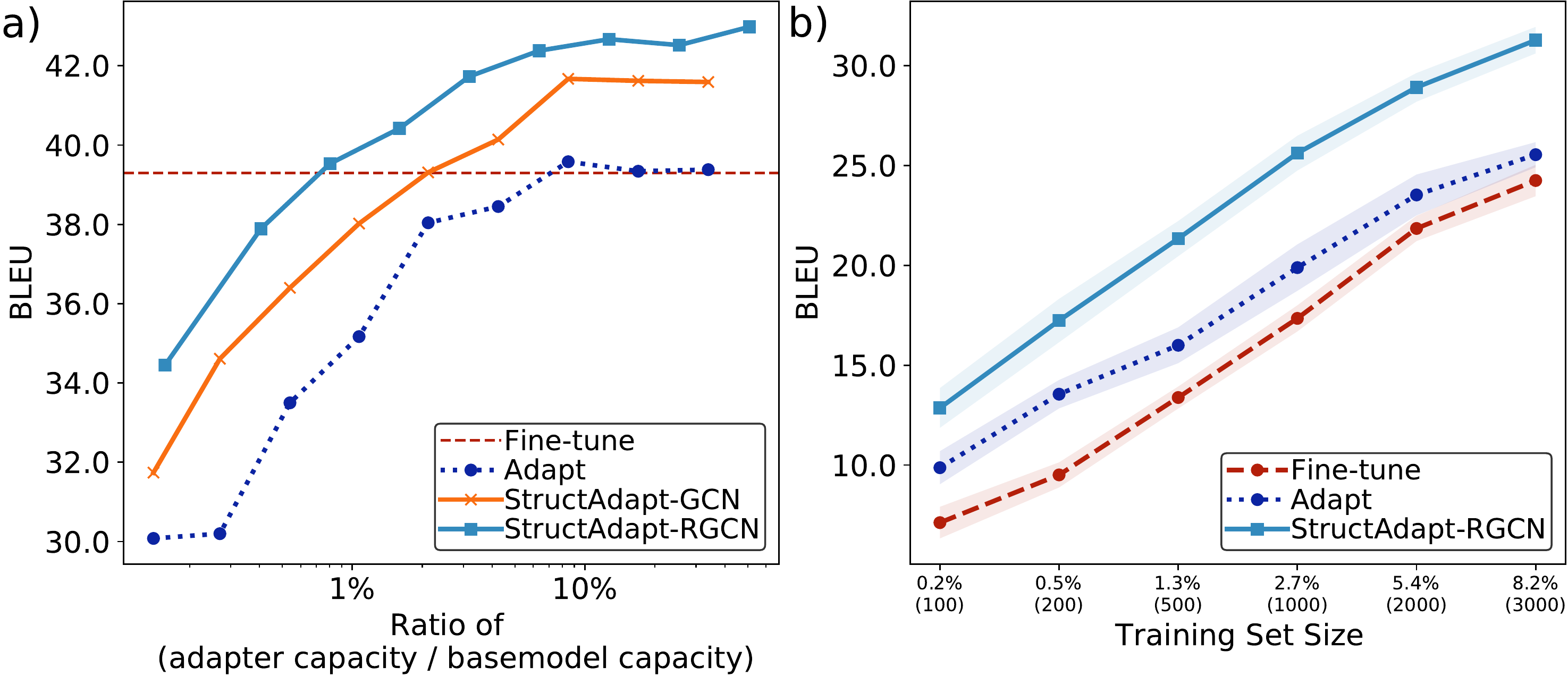}
    \caption{(a) Impact (measure with BLEU) of the number of parameters in the LDC2017T10 dev set. (b) Performance in the LDC2017T10 test set when experimenting with different amounts of training data.}
    \label{fig:hidden}
\end{figure}

\subsection{Detailed Discussion}

\paragraph{Parameter/Performance Trade-off.}

We investigate how the number of parameters affects the models. A higher hidden dimensionality means more trainable parameters, and smaller adapters introduce fewer parameters at a possible cost to performance. That is, the adapter size controls the parameter efficiency. Figure~\ref{fig:hidden}a shows the effect of the number of trained parameters in the performance measured using BLEU. Each point in the \vanilladapter and \graphadapter curves represents a hidden dimension in the range $[8, 16, \ldots, 2048]$. \graphadaptergcn is consistently better than \vanilladapter over all model capacities, even though both approaches train the same number of parameters. \graphadapterrgcn achieves similar performance than \finetune when training only 0.8\% of the parameters whereas \vanilladapter achieves similar performance to 8.5\%, demonstrating the effectiveness of injecting the graph structure into the PLM.

\paragraph{Low-data Setting.}
Previous work \cite{li2021prefixtuning} has shown that lightweight fine-tuning has an advantage in some generation tasks when the training size is smaller. Therefore, we investigate how \graphadapter behaves in a low-data setting. We subsample the LDC2017T10 training set to analyze different smaller training sets. For each size, we sample 5 different datasets and average over 2 training random seeds. Thus, we average over 10 models to get an estimate for each low-data setting.\footnote{We use the LDC2017T10 dev set to choose hyperparameters and do early stopping.} Figure~\ref{fig:hidden}b shows the results. First note that both adapter-based approaches improve over \finetune. When training with only 1000 datapoints, \graphadapter outperforms \finetune by 8.2 BLEU points. Also note that the gap between \vanilladapter and \finetune decreases when the size of the training set increases. In general, \graphadapter outperforms \finetune and \vanilladapter in low-resource scenarios by 7.3 and 4.8 BLEU points on average, respectively, whereas requiring much fewer trained parameters than \finetune and fewer number of parameters than \vanilladapter. 

\begin{table}[!t]

	\small
	\centering
	\setlength{\tabcolsep}{3pt}
	\setlength\extrarowheight{-4pt}
	\begin{tabular}{p{7.5cm}}
		\toprule
		(b / break-up-08 \\
      \quad\quad :ARG1 {\color{red}\textbf{(i / i)}} \\
      \quad\quad :ARG3 (p / person \\
            \quad\quad\quad\quad :ARG0-of (h / have-rel-role-91 \\
                  \quad\quad\quad\quad\quad\quad :ARG1 (p2 / person \\
                        \quad\quad\quad\quad\quad\quad\quad\quad:ARG0-of (h2 / have-rel-role-91 \\
                              \quad\quad\quad\quad\quad\quad\quad\quad\quad\quad:ARG1 {\color{red}\textbf{i}} \\
                              \quad\quad\quad\quad\quad\quad\quad\quad\quad\quad :ARG2 (s3 / son))) \\
                  \quad\quad\quad\quad\quad\quad :ARG2 (f / father))) \\
      \quad\quad :time (s2 / since \\
            \quad\quad\quad\quad :op1 (d / date-entity :month 8))) \\
		\midrule
		{\small\textsc{reference}}: Me and my son's father have been broken up since August.\\
		\midrule
		\finetune{\small\textsc{-2000}}: I've broken up with my son and father since August.\\
		\midrule
		\finetune: I've been with my son's father since August.\\
		\midrule
		\graphadapter{\small\textsc{-2000}}: Since August 8 I have broken up with my son's father. \\
		\midrule
		\graphadapter: I've been breaking up with my son's father since August. \\
		\bottomrule
	\end{tabular}

	\caption{An example of an AMR graph and generated sentences by different models trained on full data and on a low-data setting with 2000 datapoints.}
	\label{tab:sampleamr}
\end{table}

\paragraph{Case Study.}

We perform a case study to provide a better understanding of the \graphadapter's performance. Table~\ref{tab:sampleamr} shows an AMR graph in {\small\textsc{penman}} notation containing reentrancies (marked in bold) and sentences generated by \finetune and \graphadapter trained on the LDC2017T10 full training set and in a low-data setting where the models are trained with 2000 data points. \finetune fails in generating a sentence with the correct concept \emph{break-up} whereas \graphadapter correctly generates a sentence that describes the input graph. The incorrect verb tense is due to lack of tense information in AMR. \finetune{\small\textsc{-2000}} mixes the semantic relation between \emph{I} and \emph{son} (i.e., mistranslation of the edges in the graph) whereas \graphadapter{\small\textsc{-2000}} generates a correct sentence (except by generating the number 8).
Overall, \graphadapter produces a more accurate text output than \finetune by generating correct pronouns and mentions when control verbs and reentrancies are involved, in both full and low-data scenarios.

\begin{table}[t]
\small
\centering
{\renewcommand{\arraystretch}{0.9}

\begin{tabular}{@{\hspace*{1mm}}l@{\hspace*{4mm}}c@{\hspace*{3mm}}c@{\hspace*{1mm}}} 
\toprule
 &  \textbf{BLEU} & \textbf{BERT} \\
\midrule
\finetune & 38.5 &  95.6\\
\midrule
\vanilladapter {\small\xspace\textsc{only enc}}  & 38.5 &  95.7\\
\vanilladapter {\small\xspace\textsc{only dec}}  & 11.6 & 90.3\\
\vanilladapter {\small\xspace\textsc{enc + dec}}  & 38.6 & 95.6\\
\midrule
\graphadaptergcn {\small\xspace\textsc{only enc}}  & 40.3 & 95.9\\
\graphadaptergcn {\small\xspace\textsc{enc + dec}}  & 41.7 & 96.0\\
\bottomrule
\end{tabular}}
\caption{Impact of the adapter modules in the encoder or decoder in the LDC2017T10 dev set. All adapter-based models have the same number of parameters.}
\label{tab:adaptersparameters}
\end{table}

\paragraph{Model Variations.}

In Table~\ref{tab:adaptersparameters}, we report an ablation study on the impact of distinct adapter components, using adapters only in the encoder or decoder. We evaluate different architecture configurations keeping the same number of parameters for a fair comparison. We find that only training adapters in the decoder is not sufficient for a good performance, even having the same number of parameters. This suggests that adapting the PLM encoder to handle graph structures is key in AMR-to-text tasks. Interestingly, the model that only employs \graphadapter in the encoder (i.e., no \vanilladapter is used in the decoder) has a better performance ($+1.7$ BLEU) than using \vanilladapter in both encoder and decoder, highlighting \graphadapter's strong graph encoding abilities. Finally, the best performance is achieved when we employ \graphadapter in the encoder and \vanilladapter in the decoder, reaching 41.7 BLEU points.

\section{Graph Representation Evaluation}
\label{sec:graphrepeval}

In this section, we explore how different graph properties impact the models' abilities to encode the input graph structure.

\subsection{Impact of the Graph Representation}
\label{sec:amrinputrep}
 Inspired by~\citet{damonte-cohen-2019-structural}, we investigate two different approaches when linearizing the AMR: (i) only nodes have explicit representations, whereas edge relations are represented by the adapter parameters using the RGCN;\footnote{We use regularization based on the basis decomposition for relation weights \cite{Schlichtkrull2018ModelingRD} since AMR can contain around 150 different edge types.} and (ii) the sequence of nodes and edges using depth-first traversal of the graph. 
 
 We also propose and evaluate three different graph structures based on subwords (cf.\ \S\ref{sec:graphrep}): \emph{rep1}: for each edge, we connect every token from the source node to every token of the target node; \emph{rep2}: we connect the last token of the source node to the first token of the target node and connect the tokens of a node sequentially; \emph{rep3}: we connect the first token of the source node to the first token of the target node and connect the token of a node sequentially. Figure~\ref{fig:amrreps} shows an example of the three representations for an AMR graph structure. Additionally, we also investigate a fully connected graph structure (\emph{complete graph}), that is, similarly to the self-attention mechanism in Transformers, all nodes and edges are connected.

 \begin{table}[t]
\small
\centering
{\renewcommand{\arraystretch}{0.9}

\begin{tabular}{l@{\hspace*{1mm}}M{2.2cm}@{\hspace*{2mm}}M{0.8cm}M{0.8cm}}  
\toprule
\textbf{Linearization} & \textbf{Graph Representation} & \textbf{BLEU} & \textbf{BERT}  \\
\midrule
\multirow{ 3}{*}{(i) only nodes} & \emph{rep1} & 39.1 & 95.8\\
 & \emph{rep2} & 38.5 & 95.6\\
 & \emph{rep3} & 38.9 & 95.7\\
\midrule
\multirow{ 3}{*}{(ii) nodes and edges} & \emph{rep1} & 41.7 & 96.0\\
& \emph{rep2} & 40.4 & 95.8\\
& \emph{rep3} & 40.8 & 95.9\\
& \emph{complete graph} & 39.4 & 95.8\\

\bottomrule
\end{tabular}}
\caption{Performance on the LDC2017T10 dev set when using different graph representation strategies.}
\label{tab:graphrep}
\end{table}

As shown in Table~\ref{tab:graphrep}, explicitly considering nodes and edges in the graph linearization is beneficial. This approach has the advantage of allowing the model to handle new edge relations during inference, as they are not encoded as model parameters. Note that the \emph{complete graph} representation has relatively inferior performance, again demonstrating the advantage of explicitly encoding the input graph connectivity. 

Finally, we observe that the best configuration is using nodes and edges with \emph{rep1} (see an example in Figure~\ref{fig:amrreps}c). We believe that this is because \emph{rep1} allows direct interactions between all source and target tokens, making all token representations of an AMR node directly influenced by the neighbouring tokens.

\subsection{Robustness to Graph Linearization}
A critical advantage of modeling the graph structure is to be less dependent on linearization strategies because the graph connectivity is invariant to the graph linearization. We thus are interested in measuring the impact of the graph linearization in the models. 

Following \citet{hoyle2020promoting}, we investigate three different graph linearizations: (i)  {\small\textsc{canon}}: the original order of the canonical human-created linearizations in AMR corpora; (ii) {\small\textsc{reconf}}: the order from the canonical graph linearization is ignored, except for the top node;\footnote{{\small\textsc{reconf}} can significantly modify the linearization, including shifting edge labels (e.g., poss to poss-of).} and (iii) {\small\textsc{random}}: constructs a linearization from a random node in the graph, disregarding all order information from the canonical format, but it remains a valid traversal of the graph. All linearizations are converted to a sequence of node and edge labels using depth-first traversal and used for both training and evaluation. Examples of such graph linearizations are shown in Appendix~\ref{appe:exgraphline}. 

Table~\ref{tab:differentgraphlinearizations} presents the results. Note that while {\small\textsc{reconf}} has a negative impact on all models, \graphadapter has the best performance. \vanilladapter has similar performance gains over \finetune in all graph linearizations. Finally, note that for {\small\textsc{random}}, there is a drastic performance drop in \finetune and the gap between \graphadapter and \finetune is widest ($+5.9$ BLEU), demonstrating that explicitly encoding the graph structure is beneficial and that \graphadapter is much less impacted by different graph linearizations.

\begin{table}[t]
\small
\centering
{\renewcommand{\arraystretch}{0.9}

\begin{tabular}{l@{\hspace*{3mm}}c@{\hspace*{2.6mm}}c@{\hspace*{2.6mm}}c} 
\toprule
 & {\small\textsc{canon}} & {\small\textsc{reconf}} & {\small\textsc{random}}  \\
\midrule
\finetune & 38.0 & 35.6 & 31.3 \\
\vanilladapter  & +0.9 & +0.8 & +0.9\\
\graphadapterrgcn  & \textbf{+4.1}  & \textbf{+3.6} & \textbf{+5.9} \\
\bottomrule
\end{tabular}}
\caption{Differences, with respect to \finetune, in the BLEU score of the LDC2017T10 test set as a function of different graph linearizations.}
\label{tab:differentgraphlinearizations}
\end{table}

\subsection{Graph Properties}

Table~\ref{tab:reentrances} shows the effects of the graph size, graph diameter and reentrancies in the performance. First, note that the BLEU scores decrease as the graph size increases since larger graphs often are more complex. The performance gap between \graphadapter and \finetune becomes larger for relatively larger graphs, showing that \graphadapter is able to better encode complex graphs. As \vanilladapter is not aware of the graph connectivity, it has much worse scores compared to \graphadapter, especially for larger graphs. 

It is expected that the beneﬁt of the \graphadapter will be more evident for AMR graphs containing larger diameter as the encoder is aware of the input graph structure. As seen in Table~\ref{tab:reentrances}, similarly to the graph size, the scores decrease as the graph diameter increases. \graphadapter achieves a clear improvement when handling graphs with ${\geq}20$ diameter, with a improvement of $+4.2$ BLEU points over \finetune. 

Previous work \cite{damonte-cohen-2019-structural,szubert-etal-2020-role} showed that reentrancies (nodes with multiple parents) pose difficulties in encoding AMRs correctly. Because \graphadapter is the only approach to model reentrancies explicitly, we expect it to deal better with these structures. The gap between \graphadapter and the other models is widest for examples with more reentrancies, confirming our hypothesis. In particular, when graphs contain ${\geq}4$ reentrancies, \graphadapter has an improvement of $+3.6$ BLEU points compared to \vanilladapter.

\begin{table}[t]
\small
\centering
{\renewcommand{\arraystretch}{0.9}

\begin{tabular}{lccc} 
\toprule
 \textbf{graph size}& 1-30 & 31-60 & ${>}60$  \\
 \textbf{$\#$ datapoints} & 548 & 537 & 286  \\
\midrule
\finetune & 40.6 & 37.3 & 38.1 \\
\vanilladapter  & +0.5 & +1.4 & +1.1 \\
\graphadapterrgcn  & \textbf{+2.3}  & \textbf{+4.0} & \textbf{+4.6} \\
\toprule
 \textbf{graph diameter}& 1-10 & 11-20 & ${>}20$  \\
 \textbf{$\#$ datapoints} & 384 & 769 & 218  \\
\midrule
\finetune & 43.3 & 37.6 & 38.5  \\
\vanilladapter  & -0.1 & +1.7 & +0.3 \\
\graphadapterrgcn  & \textbf{+0.5}  & \textbf{+4.3} & \textbf{+4.2} \\
\toprule
 \textbf{$\#$ reentrancies}& 0 & 1-3 & 4-20  \\
 \textbf{$\#$ datapoints} & 619 & 664 & 88  \\
\midrule
\finetune & 42.9 & 38.0 & 31.3 \\
\vanilladapter  & +0.2 & +1.7 & +0.8 \\
\graphadapterrgcn  & \textbf{+3.4}  & \textbf{+4.4} & \textbf{+4.4} \\
\bottomrule
\end{tabular}}
\caption{Differences, with respect to  \finetune, in the BLEU score of the LDC2017T10 test set as a function of the graph size, graph diameter and number of reentrancies.}
\label{tab:reentrances}

\end{table}

\section{Conclusion}

We presented \graphadapter, a novel adapter architecture to explicitly model graph structures into pretrained language models, providing an extensive evaluation of our approach and showing that it achieves state-of-the-art results on two AMR-to-text benchmarks, training much fewer parameters. We also found that \graphadapter is more effective when encoding complex graphs, when trained on fewer datapoints, and is more robust to different graph linearizations and reentrancies. In future work, we plan to consider other graph-to-text tasks, such as those based on Knowledge Graphs.

\section*{Acknowledgments}
We thank our anonymous reviewers for their thoughtful comments. We also would like to thank Jonas Pfeiffer, Jorge Cardona, Juri Opitz, Kevin Stowe, Thy Tran, Tilman Beck and Tim Baumg{\"a}rtner for their feedback on this work. This work has been supported by the German Research Foundation (DFG) as part of the Research Training Group ``Adaptive Preparation of Information form Heterogeneous Sources'' (AIPHES, GRK 1994/1) and as part of the DFG funded project UKP-SQuARE with the number GU 798/29-1. 

\bibliography{anthology,custom}

\begin{thebibliography}{65}
\expandafter\ifx\csname natexlab\endcsname\relax\def\natexlab#1{#1}\fi

\bibitem[{Bai et~al.(2020)Bai, Song, and Zhang}]{bai-etal-2020-online}
Xuefeng Bai, Linfeng Song, and Yue Zhang. 2020.
\newblock \href {https://doi.org/10.18653/v1/2020.emnlp-main.92} {Online
  back-parsing for {AMR}-to-text generation}.
\newblock In \emph{Proceedings of the 2020 Conference on Empirical Methods in
  Natural Language Processing (EMNLP)}, pages 1206--1219, Online. Association
  for Computational Linguistics.

\bibitem[{Banarescu et~al.(2013)Banarescu, Bonial, Cai, Georgescu, Griffitt,
  Hermjakob, Knight, Koehn, Palmer, and
  Schneider}]{banarescu-etal-2013-abstract}
Laura Banarescu, Claire Bonial, Shu Cai, Madalina Georgescu, Kira Griffitt, Ulf
  Hermjakob, Kevin Knight, Philipp Koehn, Martha Palmer, and Nathan Schneider.
  2013.
\newblock \href {https://www.aclweb.org/anthology/W13-2322} {{A}bstract
  {M}eaning {R}epresentation for sembanking}.
\newblock In \emph{Proceedings of the 7th Linguistic Annotation Workshop and
  Interoperability with Discourse}, pages 178--186, Sofia, Bulgaria.
  Association for Computational Linguistics.

\bibitem[{Beck et~al.(2018)Beck, Haffari, and Cohn}]{beck-etal-2018-graph}
Daniel Beck, Gholamreza Haffari, and Trevor Cohn. 2018.
\newblock \href {https://doi.org/10.18653/v1/P18-1026} {Graph-to-sequence
  learning using gated graph neural networks}.
\newblock In \emph{Proceedings of the 56th Annual Meeting of the Association
  for Computational Linguistics (Volume 1: Long Papers)}, pages 273--283,
  Melbourne, Australia. Association for Computational Linguistics.

\bibitem[{Bevilacqua et~al.(2021)Bevilacqua, Blloshmi, and
  Navigli}]{Micheleamr}
Michele Bevilacqua, Rexhina Blloshmi, and Roberto Navigli. 2021.
\newblock \href {https://ojs.aaai.org/index.php/AAAI/article/view/17489} {One
  spring to rule them both: Symmetric amr semantic parsing and generation
  without a complex pipeline}.
\newblock \emph{Proceedings of the AAAI Conference on Artificial Intelligence},
  35(14):12564--12573.

\bibitem[{Cai and Lam(2020)}]{cai-lam-2020-graph}
Deng Cai and Wai Lam. 2020.
\newblock \href {https://doi.org/10.1609/aaai.v34i05.6243} {Graph transformer
  for graph-to-sequence learning}.
\newblock \emph{Proceedings of the AAAI Conference on Artificial Intelligence},
  34(05):7464--7471.

\bibitem[{Castro~Ferreira et~al.(2019)Castro~Ferreira, van~der Lee, van
  Miltenburg, and Krahmer}]{castro-ferreira-etal-2019-neural}
Thiago Castro~Ferreira, Chris van~der Lee, Emiel van Miltenburg, and Emiel
  Krahmer. 2019.
\newblock \href {https://doi.org/10.18653/v1/D19-1052} {Neural data-to-text
  generation: A comparison between pipeline and end-to-end architectures}.
\newblock In \emph{Proceedings of the 2019 Conference on Empirical Methods in
  Natural Language Processing and the 9th International Joint Conference on
  Natural Language Processing (EMNLP-IJCNLP)}, pages 552--562, Hong Kong,
  China. Association for Computational Linguistics.

\bibitem[{Damonte and Cohen(2019)}]{damonte-cohen-2019-structural}
Marco Damonte and Shay~B. Cohen. 2019.
\newblock \href {https://doi.org/10.18653/v1/N19-1366} {Structural neural
  encoders for {AMR}-to-text generation}.
\newblock In \emph{Proceedings of the 2019 Conference of the North {A}merican
  Chapter of the Association for Computational Linguistics: Human Language
  Technologies, Volume 1 (Long and Short Papers)}, pages 3649--3658,
  Minneapolis, Minnesota. Association for Computational Linguistics.

\bibitem[{Devlin et~al.(2019)Devlin, Chang, Lee, and
  Toutanova}]{devlin-etal-2019-bert}
Jacob Devlin, Ming-Wei Chang, Kenton Lee, and Kristina Toutanova. 2019.
\newblock \href {https://doi.org/10.18653/v1/N19-1423} {{BERT}: Pre-training of
  deep bidirectional transformers for language understanding}.
\newblock In \emph{Proceedings of the 2019 Conference of the North {A}merican
  Chapter of the Association for Computational Linguistics: Human Language
  Technologies, Volume 1 (Long and Short Papers)}, pages 4171--4186,
  Minneapolis, Minnesota. Association for Computational Linguistics.

\bibitem[{Flanigan et~al.(2016)Flanigan, Dyer, Smith, and
  Carbonell}]{flanigan-etal-2016-generation}
Jeffrey Flanigan, Chris Dyer, Noah~A. Smith, and Jaime Carbonell. 2016.
\newblock \href {https://doi.org/10.18653/v1/N16-1087} {Generation from
  {A}bstract {M}eaning {R}epresentation using tree transducers}.
\newblock In \emph{Proceedings of the 2016 Conference of the North {A}merican
  Chapter of the Association for Computational Linguistics: Human Language
  Technologies}, pages 731--739, San Diego, California. Association for
  Computational Linguistics.

\bibitem[{Fu et~al.(2021)Fu, Song, Du, and Zhang}]{fu-etal-2021-end}
Qiankun Fu, Linfeng Song, Wenyu Du, and Yue Zhang. 2021.
\newblock \href {https://doi.org/10.18653/v1/2021.acl-long.324} {End-to-end
  {AMR} corefencence resolution}.
\newblock In \emph{Proceedings of the 59th Annual Meeting of the Association
  for Computational Linguistics and the 11th International Joint Conference on
  Natural Language Processing (Volume 1: Long Papers)}, pages 4204--4214,
  Online. Association for Computational Linguistics.

\bibitem[{Gardent et~al.(2017)Gardent, Shimorina, Narayan, and
  Perez-Beltrachini}]{gardent-etal-2017-webnlg}
Claire Gardent, Anastasia Shimorina, Shashi Narayan, and Laura
  Perez-Beltrachini. 2017.
\newblock \href {https://doi.org/10.18653/v1/W17-3518} {The {W}eb{NLG}
  challenge: Generating text from {RDF} data}.
\newblock In \emph{Proceedings of the 10th International Conference on Natural
  Language Generation}, pages 124--133, Santiago de Compostela, Spain.
  Association for Computational Linguistics.

\bibitem[{Goodfellow et~al.(2014)Goodfellow, Mirza, Xiao, Courville, and
  Bengio}]{goodfellow2013an}
Ian~J. Goodfellow, Mehdi Mirza, Da~Xiao, Aaron Courville, and Yoshua Bengio.
  2014.
\newblock \href {https://openreview.net/forum?id=oXSw7laxwUpln} {An empirical
  investigation of catastrophic forgeting in gradient-based neural networks}.
\newblock In \emph{Proceedings of International Conference on Learning
  Representations (ICLR)}.

\bibitem[{Guo et~al.(2019)Guo, Zhang, Teng, and Lu}]{doi:10.116200269}
Zhijiang Guo, Yan Zhang, Zhiyang Teng, and Wei Lu. 2019.
\newblock \href {https://doi.org/10.1162/tacl\_a\_00269} {Densely connected
  graph convolutional networks for graph-to-sequence learning}.
\newblock \emph{Transactions of the Association for Computational Linguistics},
  7:297--312.

\bibitem[{Hambardzumyan et~al.(2021)Hambardzumyan, Khachatrian, and
  May}]{hambardzumyan-etal-2021-warp}
Karen Hambardzumyan, Hrant Khachatrian, and Jonathan May. 2021.
\newblock \href {https://doi.org/10.18653/v1/2021.acl-long.381} {{WARP}:
  {W}ord-level {A}dversarial {R}e{P}rogramming}.
\newblock In \emph{Proceedings of the 59th Annual Meeting of the Association
  for Computational Linguistics and the 11th International Joint Conference on
  Natural Language Processing (Volume 1: Long Papers)}, pages 4921--4933,
  Online. Association for Computational Linguistics.

\bibitem[{Harkous et~al.(2020)Harkous, Groves, and Saffari}]{harkous2020text}
Hamza Harkous, Isabel Groves, and Amir Saffari. 2020.
\newblock \href {https://doi.org/10.18653/v1/2020.coling-main.218} {Have your
  text and use it too! end-to-end neural data-to-text generation with semantic
  fidelity}.
\newblock In \emph{Proceedings of the 28th International Conference on
  Computational Linguistics}, pages 2410--2424, Barcelona, Spain (Online).
  International Committee on Computational Linguistics.

\bibitem[{Houlsby et~al.(2019)Houlsby, Giurgiu, Jastrzebski, Morrone,
  De~Laroussilhe, Gesmundo, Attariyan, and Gelly}]{pmlr-v97-houlsby19a}
Neil Houlsby, Andrei Giurgiu, Stanislaw Jastrzebski, Bruna Morrone, Quentin
  De~Laroussilhe, Andrea Gesmundo, Mona Attariyan, and Sylvain Gelly. 2019.
\newblock \href {http://proceedings.mlr.press/v97/houlsby19a.html}
  {Parameter-efficient transfer learning for {NLP}}.
\newblock In \emph{Proceedings of the 36th International Conference on Machine
  Learning}, volume~97 of \emph{Proceedings of Machine Learning Research},
  pages 2790--2799. PMLR.

\bibitem[{Hovy et~al.(2013)Hovy, Berg-Kirkpatrick, Vaswani, and
  Hovy}]{hovy-etal-2013-learning}
Dirk Hovy, Taylor Berg-Kirkpatrick, Ashish Vaswani, and Eduard Hovy. 2013.
\newblock \href {https://www.aclweb.org/anthology/N13-1132} {Learning whom to
  trust with {MACE}}.
\newblock In \emph{Proceedings of the 2013 Conference of the North {A}merican
  Chapter of the Association for Computational Linguistics: Human Language
  Technologies}, pages 1120--1130, Atlanta, Georgia. Association for
  Computational Linguistics.

\bibitem[{Howard and Ruder(2018)}]{howard-ruder-2018-universal}
Jeremy Howard and Sebastian Ruder. 2018.
\newblock \href {https://doi.org/10.18653/v1/P18-1031} {Universal language
  model fine-tuning for text classification}.
\newblock In \emph{Proceedings of the 56th Annual Meeting of the Association
  for Computational Linguistics (Volume 1: Long Papers)}, pages 328--339,
  Melbourne, Australia. Association for Computational Linguistics.

\bibitem[{Hoyle et~al.(2021)Hoyle, Marasovi{\'c}, and
  Smith}]{hoyle2020promoting}
Alexander~Miserlis Hoyle, Ana Marasovi{\'c}, and Noah~A. Smith. 2021.
\newblock \href {https://doi.org/10.18653/v1/2021.findings-acl.82} {Promoting
  graph awareness in linearized graph-to-text generation}.
\newblock In \emph{Findings of the Association for Computational Linguistics:
  ACL-IJCNLP 2021}, pages 944--956, Online. Association for Computational
  Linguistics.

\bibitem[{Kale(2020)}]{kale2020texttotext}
Mihir Kale. 2020.
\newblock \href {http://arxiv.org/abs/2005.10433} {Text-to-text pre-training
  for data-to-text tasks}.
\newblock \emph{arXiv e-prints}.

\bibitem[{Kingma and Ba(2015)}]{kingma:adam}
Diederik~P. Kingma and Jimmy Ba. 2015.
\newblock \href {http://arxiv.org/abs/1412.6980} {Adam: {A} method for
  stochastic optimization}.
\newblock In \emph{3rd International Conference on Learning Representations,
  {ICLR} 2015, San Diego, CA, USA, May 7-9, 2015, Conference Track
  Proceedings}.

\bibitem[{Kipf and Welling(2017)}]{Kipf:2016tc}
Thomas~N. Kipf and Max Welling. 2017.
\newblock \href {https://openreview.net/forum?id=SJU4ayYgl} {Semi-supervised
  classification with graph convolutional networks}.
\newblock In \emph{5th International Conference on Learning Representations,
  {ICLR} 2017, Toulon, France, April 24-26, 2017, Conference Track
  Proceedings}.

\bibitem[{Kirkpatrick et~al.(2017)Kirkpatrick, Pascanu, Rabinowitz, Veness,
  Desjardins, Rusu, Milan, Quan, Ramalho, Grabska-Barwinska, Hassabis, Clopath,
  Kumaran, and Hadsell}]{Kirkpatrick3521}
James Kirkpatrick, Razvan Pascanu, Neil Rabinowitz, Joel Veness, Guillaume
  Desjardins, Andrei~A. Rusu, Kieran Milan, John Quan, Tiago Ramalho, Agnieszka
  Grabska-Barwinska, Demis Hassabis, Claudia Clopath, Dharshan Kumaran, and
  Raia Hadsell. 2017.
\newblock \href {https://doi.org/10.1073/pnas.1611835114} {Overcoming
  catastrophic forgetting in neural networks}.
\newblock \emph{Proceedings of the National Academy of Sciences},
  114(13):3521--3526.

\bibitem[{Konstas et~al.(2017)Konstas, Iyer, Yatskar, Choi, and
  Zettlemoyer}]{konstas-etal-2017-neural}
Ioannis Konstas, Srinivasan Iyer, Mark Yatskar, Yejin Choi, and Luke
  Zettlemoyer. 2017.
\newblock \href {https://doi.org/10.18653/v1/P17-1014} {Neural {AMR}:
  Sequence-to-sequence models for parsing and generation}.
\newblock In \emph{Proceedings of the 55th Annual Meeting of the Association
  for Computational Linguistics (Volume 1: Long Papers)}, pages 146--157,
  Vancouver, Canada. Association for Computational Linguistics.

\bibitem[{Lauscher et~al.(2020)Lauscher, Majewska, Ribeiro, Gurevych, Rozanov,
  and Glava{\v{s}}}]{lauscher-etal-2020-common}
Anne Lauscher, Olga Majewska, Leonardo F.~R. Ribeiro, Iryna Gurevych, Nikolai
  Rozanov, and Goran Glava{\v{s}}. 2020.
\newblock \href {https://doi.org/10.18653/v1/2020.deelio-1.5} {Common sense or
  world knowledge? investigating adapter-based knowledge injection into
  pretrained transformers}.
\newblock In \emph{Proceedings of Deep Learning Inside Out (DeeLIO): The First
  Workshop on Knowledge Extraction and Integration for Deep Learning
  Architectures}, pages 43--49, Online. Association for Computational
  Linguistics.

\bibitem[{Lewis et~al.(2020)Lewis, Liu, Goyal, Ghazvininejad, Mohamed, Levy,
  Stoyanov, and Zettlemoyer}]{lewis2019bart}
Mike Lewis, Yinhan Liu, Naman Goyal, Marjan Ghazvininejad, Abdelrahman Mohamed,
  Omer Levy, Veselin Stoyanov, and Luke Zettlemoyer. 2020.
\newblock \href {https://doi.org/10.18653/v1/2020.acl-main.703} {{BART}:
  Denoising sequence-to-sequence pre-training for natural language generation,
  translation, and comprehension}.
\newblock In \emph{Proceedings of the 58th Annual Meeting of the Association
  for Computational Linguistics}, pages 7871--7880, Online. Association for
  Computational Linguistics.

\bibitem[{Li and Liang(2021)}]{li2021prefixtuning}
Xiang~Lisa Li and Percy Liang. 2021.
\newblock \href {https://doi.org/10.18653/v1/2021.acl-long.353} {Prefix-tuning:
  Optimizing continuous prompts for generation}.
\newblock In \emph{Proceedings of the 59th Annual Meeting of the Association
  for Computational Linguistics and the 11th International Joint Conference on
  Natural Language Processing (Volume 1: Long Papers)}, pages 4582--4597,
  Online. Association for Computational Linguistics.

\bibitem[{Liao et~al.(2018)Liao, Lebanoff, and Liu}]{liao-etal-2018-abstract}
Kexin Liao, Logan Lebanoff, and Fei Liu. 2018.
\newblock \href {https://www.aclweb.org/anthology/C18-1101} {{A}bstract
  {M}eaning {R}epresentation for multi-document summarization}.
\newblock In \emph{Proceedings of the 27th International Conference on
  Computational Linguistics}, pages 1178--1190, Santa Fe, New Mexico, USA.
  Association for Computational Linguistics.

\bibitem[{Liu et~al.(2021)Liu, Zheng, Du, Ding, Qian, Yang, and
  Tang}]{DBLP:journals/corr/abs-2103-10385}
Xiao Liu, Yanan Zheng, Zhengxiao Du, Ming Ding, Yujie Qian, Zhilin Yang, and
  Jie Tang. 2021.
\newblock \href {http://arxiv.org/abs/2103.10385} {{GPT} understands, too}.
\newblock \emph{CoRR}, abs/2103.10385.

\bibitem[{Liu et~al.(2020)Liu, Ott, Goyal, Du, Joshi, Chen, Levy, Lewis,
  Zettlemoyer, and Stoyanov}]{liu2020roberta}
Yinhan Liu, Myle Ott, Naman Goyal, Jingfei Du, Mandar Joshi, Danqi Chen, Omer
  Levy, Mike Lewis, Luke Zettlemoyer, and Veselin Stoyanov. 2020.
\newblock \href {https://openreview.net/forum?id=SyxS0T4tvS} {Roberta: A
  robustly optimized bert pretraining approach}.
\newblock \emph{arXiv e-prints}.

\bibitem[{Mager et~al.(2020)Mager, Fernandez~Astudillo, Naseem, Sultan, Lee,
  Florian, and Roukos}]{mager2020gpttoo}
Manuel Mager, Ram{\'o}n Fernandez~Astudillo, Tahira Naseem, Md~Arafat Sultan,
  Young-Suk Lee, Radu Florian, and Salim Roukos. 2020.
\newblock \href {https://doi.org/10.18653/v1/2020.acl-main.167} {{GPT}-too: A
  language-model-first approach for {AMR}-to-text generation}.
\newblock In \emph{Proceedings of the 58th Annual Meeting of the Association
  for Computational Linguistics}, pages 1846--1852, Online. Association for
  Computational Linguistics.

\bibitem[{Nan et~al.(2021)Nan, Radev, Zhang, Rau, Sivaprasad, Hsieh, Tang,
  Vyas, Verma, Krishna, Liu, Irwanto, Pan, Rahman, Zaidi, Mutuma, Tarabar,
  Gupta, Yu, Tan, Lin, Xiong, Socher, and Rajani}]{radev2020dart}
Linyong Nan, Dragomir Radev, Rui Zhang, Amrit Rau, Abhinand Sivaprasad,
  Chiachun Hsieh, Xiangru Tang, Aadit Vyas, Neha Verma, Pranav Krishna,
  Yangxiaokang Liu, Nadia Irwanto, Jessica Pan, Faiaz Rahman, Ahmad Zaidi,
  Mutethia Mutuma, Yasin Tarabar, Ankit Gupta, Tao Yu, Yi~Chern Tan,
  Xi~Victoria Lin, Caiming Xiong, Richard Socher, and Nazneen~Fatema Rajani.
  2021.
\newblock \href {https://doi.org/10.18653/v1/2021.naacl-main.37} {{DART}:
  Open-domain structured data record to text generation}.
\newblock In \emph{Proceedings of the 2021 Conference of the North American
  Chapter of the Association for Computational Linguistics: Human Language
  Technologies}, pages 432--447, Online. Association for Computational
  Linguistics.

\bibitem[{Opitz et~al.(2021)Opitz, Daza, and Frank}]{bamboo}
Juri Opitz, Angel Daza, and Anette Frank. 2021.
\newblock \href {http://arxiv.org/abs/2108.11949} {Weisfeiler-leman in the
  bamboo: Novel amr graph metrics and a benchmark for amr graph similarity}.
\newblock \emph{Transactions of the Association for Computational Linguistics}.

\bibitem[{Opitz and Frank(2021)}]{opitz-frank-2021-towards}
Juri Opitz and Anette Frank. 2021.
\newblock \href {https://www.aclweb.org/anthology/2021.eacl-main.129} {Towards
  a decomposable metric for explainable evaluation of text generation from
  {AMR}}.
\newblock In \emph{Proceedings of the 16th Conference of the European Chapter
  of the Association for Computational Linguistics: Main Volume}, pages
  1504--1518, Online. Association for Computational Linguistics.

\bibitem[{Opitz et~al.(2020)Opitz, Parcalabescu, and
  Frank}]{opitz-etal-2020-amr}
Juri Opitz, Letitia Parcalabescu, and Anette Frank. 2020.
\newblock \href {https://doi.org/10.1162/tacl_a_00329} {{AMR} similarity
  metrics from principles}.
\newblock \emph{Transactions of the Association for Computational Linguistics},
  8:522--538.

\bibitem[{Papineni et~al.(2002)Papineni, Roukos, Ward, and
  Zhu}]{papineni-etal-2002-bleu}
Kishore Papineni, Salim Roukos, Todd Ward, and Wei-Jing Zhu. 2002.
\newblock \href {https://doi.org/10.3115/1073083.1073135} {{B}leu: a method for
  automatic evaluation of machine translation}.
\newblock In \emph{Proceedings of the 40th Annual Meeting of the Association
  for Computational Linguistics}, pages 311--318, Philadelphia, Pennsylvania,
  USA. Association for Computational Linguistics.

\bibitem[{Parikh et~al.(2020)Parikh, Wang, Gehrmann, Faruqui, Dhingra, Yang,
  and Das}]{parikh-etal-2020-totto}
Ankur Parikh, Xuezhi Wang, Sebastian Gehrmann, Manaal Faruqui, Bhuwan Dhingra,
  Diyi Yang, and Dipanjan Das. 2020.
\newblock \href {https://doi.org/10.18653/v1/2020.emnlp-main.89} {{ToTTo}: A
  controlled table-to-text generation dataset}.
\newblock In \emph{Proceedings of the 2020 Conference on Empirical Methods in
  Natural Language Processing (EMNLP)}, pages 1173--1186, Online. Association
  for Computational Linguistics.

\bibitem[{Pfeiffer et~al.(2021)Pfeiffer, Kamath, Rücklé, Cho, and
  Gurevych}]{pfeiffer2021adapterfusion}
Jonas Pfeiffer, Aishwarya Kamath, Andreas Rücklé, Kyunghyun Cho, and Iryna
  Gurevych. 2021.
\newblock \href {https://arxiv.org/abs/2005.00247} {{AdapterFusion:
  Non-Destructive Task Composition for Transfer Learning}}.
\newblock In \emph{Proceedings of the 16th Conference of the {E}uropean Chapter
  of the Association for Computational Linguistics (EACL)}, Online. Association
  for Computational Linguistics.

\bibitem[{Pfeiffer et~al.(2020{\natexlab{a}})Pfeiffer, R{\"u}ckl{\'e}, Poth,
  Kamath, Vuli{\'c}, Ruder, Cho, and Gurevych}]{pfeiffer-etal-2020-adapterhub}
Jonas Pfeiffer, Andreas R{\"u}ckl{\'e}, Clifton Poth, Aishwarya Kamath, Ivan
  Vuli{\'c}, Sebastian Ruder, Kyunghyun Cho, and Iryna Gurevych.
  2020{\natexlab{a}}.
\newblock \href {https://doi.org/10.18653/v1/2020.emnlp-demos.7}
  {{A}dapter{H}ub: A framework for adapting transformers}.
\newblock In \emph{Proceedings of the 2020 Conference on Empirical Methods in
  Natural Language Processing: System Demonstrations}, pages 46--54, Online.
  Association for Computational Linguistics.

\bibitem[{Pfeiffer et~al.(2020{\natexlab{b}})Pfeiffer, Vuli{\'c}, Gurevych, and
  Ruder}]{pfeiffer-etal-2020-mad}
Jonas Pfeiffer, Ivan Vuli{\'c}, Iryna Gurevych, and Sebastian Ruder.
  2020{\natexlab{b}}.
\newblock \href {https://doi.org/10.18653/v1/2020.emnlp-main.617} {{MAD-X}:
  {A}n {A}dapter-{B}ased {F}ramework for {M}ulti-{T}ask {C}ross-{L}ingual
  {T}ransfer}.
\newblock In \emph{Proceedings of the 2020 Conference on Empirical Methods in
  Natural Language Processing (EMNLP)}, pages 7654--7673, Online. Association
  for Computational Linguistics.

\bibitem[{Popovi{\'c}(2015)}]{popovic-2015-chrf}
Maja Popovi{\'c}. 2015.
\newblock \href {https://doi.org/10.18653/v1/W15-3049} {chr{F}: character
  n-gram {F}-score for automatic {MT} evaluation}.
\newblock In \emph{Proceedings of the Tenth Workshop on Statistical Machine
  Translation}, pages 392--395, Lisbon, Portugal. Association for Computational
  Linguistics.

\bibitem[{Pourdamghani et~al.(2016)Pourdamghani, Knight, and
  Hermjakob}]{pourdamghani-etal-2016-generating}
Nima Pourdamghani, Kevin Knight, and Ulf Hermjakob. 2016.
\newblock \href {https://doi.org/10.18653/v1/W16-6603} {Generating {E}nglish
  from {A}bstract {M}eaning {R}epresentations}.
\newblock In \emph{Proceedings of the 9th International Natural Language
  Generation conference}, pages 21--25, Edinburgh, UK. Association for
  Computational Linguistics.

\bibitem[{Radford et~al.(2019)Radford, Wu, Child, Luan, Amodei, and
  Sutskever}]{radford2019language}
Alec Radford, Jeffrey Wu, Rewon Child, David Luan, Dario Amodei, and Ilya
  Sutskever. 2019.
\newblock Language models are unsupervised multitask learners.
\newblock In \emph{Technical report, OpenAI}.

\bibitem[{Raffel et~al.(2019)Raffel, Shazeer, Roberts, Lee, Narang, Matena,
  Zhou, Li, and Liu}]{2019t5}
Colin Raffel, Noam Shazeer, Adam Roberts, Katherine Lee, Sharan Narang, Michael
  Matena, Yanqi Zhou, Wei Li, and Peter~J. Liu. 2019.
\newblock \href {http://arxiv.org/abs/1910.10683} {Exploring the limits of
  transfer learning with a unified text-to-text transformer}.
\newblock \emph{arXiv e-prints}.

\bibitem[{Rebuffi et~al.(2017)Rebuffi, Bilen, and Vedaldi}]{NIPS2017_e7b24b11}
Sylvestre-Alvise Rebuffi, Hakan Bilen, and Andrea Vedaldi. 2017.
\newblock \href
  {https://proceedings.neurips.cc/paper/2017/file/e7b24b112a44fdd9ee93bdf998c6ca0e-Paper.pdf}
  {Learning multiple visual domains with residual adapters}.
\newblock In \emph{Advances in Neural Information Processing Systems},
  volume~30, pages 506--516. Curran Associates, Inc.

\bibitem[{Ribeiro et~al.(2019)Ribeiro, Gardent, and
  Gurevych}]{ribeiro-etal-2019-enhancing}
Leonardo F.~R. Ribeiro, Claire Gardent, and Iryna Gurevych. 2019.
\newblock \href {https://doi.org/10.18653/v1/D19-1314} {Enhancing {AMR}-to-text
  generation with dual graph representations}.
\newblock In \emph{Proceedings of the 2019 Conference on Empirical Methods in
  Natural Language Processing and the 9th International Joint Conference on
  Natural Language Processing (EMNLP-IJCNLP)}, pages 3183--3194, Hong Kong,
  China. Association for Computational Linguistics.

\bibitem[{Ribeiro et~al.(2021)Ribeiro, Pfeiffer, Zhang, and
  Gurevych}]{ribeiro2021smelting}
Leonardo F.~R. Ribeiro, Jonas Pfeiffer, Yue Zhang, and Iryna Gurevych. 2021.
\newblock Smelting gold and silver for improved multilingual amr-to-text
  generation.
\newblock In \emph{Proceedings of the 2021 Conference on Empirical Methods in
  Natural Language Processing, {EMNLP} 2021, Punta Cana, November 7-11, 2021}.

\bibitem[{Ribeiro et~al.(2020{\natexlab{a}})Ribeiro, Schmitt, Schütze, and
  Gurevych}]{ribeiro2020investigating}
Leonardo F.~R. Ribeiro, Martin Schmitt, Hinrich Schütze, and Iryna Gurevych.
  2020{\natexlab{a}}.
\newblock \href {http://arxiv.org/abs/2007.08426} {Investigating pretrained
  language models for graph-to-text generation}.
\newblock \emph{arXiv e-prints}.

\bibitem[{Ribeiro et~al.(2020{\natexlab{b}})Ribeiro, Zhang, Gardent, and
  Gurevych}]{ribeiro2020modeling}
Leonardo F.~R. Ribeiro, Yue Zhang, Claire Gardent, and Iryna Gurevych.
  2020{\natexlab{b}}.
\newblock \href {https://doi.org/10.1162/tacl_a_00332} {Modeling global and
  local node contexts for text generation from knowledge graphs}.
\newblock \emph{Transactions of the Association for Computational Linguistics},
  8:589--604.

\bibitem[{Schlichtkrull et~al.(2018)Schlichtkrull, Kipf, Bloem, van~den Berg,
  Titov, and Welling}]{Schlichtkrull2018ModelingRD}
Michael~Sejr Schlichtkrull, Thomas~N. Kipf, Peter Bloem, Rianne van~den Berg,
  Ivan Titov, and Max Welling. 2018.
\newblock \href
  {https://link.springer.com/chapter/10.1007/978-3-319-93417-4_38} {Modeling
  relational data with graph convolutional networks}.
\newblock In \emph{{ESWC} 2018, Heraklion, Crete, Greece, June 3-7, 2018,
  Proceedings}, pages 593--607.

\bibitem[{Schmitt et~al.(2021)Schmitt, Ribeiro, Dufter, Gurevych, and
  Sch{\"u}tze}]{schmitt2020modeling}
Martin Schmitt, Leonardo F.~R. Ribeiro, Philipp Dufter, Iryna Gurevych, and
  Hinrich Sch{\"u}tze. 2021.
\newblock \href {https://aclanthology.org/2021.textgraphs-1.2} {Modeling graph
  structure via relative position for text generation from knowledge graphs}.
\newblock In \emph{Proceedings of the Fifteenth Workshop on Graph-Based Methods
  for Natural Language Processing (TextGraphs-15)}, pages 10--21, Mexico City,
  Mexico. Association for Computational Linguistics.

\bibitem[{Sennrich et~al.(2016)Sennrich, Haddow, and
  Birch}]{sennrich-etal-2016-neural}
Rico Sennrich, Barry Haddow, and Alexandra Birch. 2016.
\newblock \href {https://doi.org/10.18653/v1/P16-1162} {Neural machine
  translation of rare words with subword units}.
\newblock In \emph{Proceedings of the 54th Annual Meeting of the Association
  for Computational Linguistics (Volume 1: Long Papers)}, pages 1715--1725,
  Berlin, Germany. Association for Computational Linguistics.

\bibitem[{Song et~al.(2019)Song, Gildea, Zhang, Wang, and
  Su}]{doi:10.116200252}
Linfeng Song, Daniel Gildea, Yue Zhang, Zhiguo Wang, and Jinsong Su. 2019.
\newblock \href {https://doi.org/10.1162/tacl\_a\_00252} {Semantic neural
  machine translation using amr}.
\newblock \emph{Transactions of the Association for Computational Linguistics},
  7:19--31.

\bibitem[{Song et~al.(2018)Song, Zhang, Wang, and
  Gildea}]{song-etal-2018-graph}
Linfeng Song, Yue Zhang, Zhiguo Wang, and Daniel Gildea. 2018.
\newblock \href {https://doi.org/10.18653/v1/P18-1150} {A graph-to-sequence
  model for {AMR}-to-text generation}.
\newblock In \emph{Proceedings of the 56th Annual Meeting of the Association
  for Computational Linguistics (Volume 1: Long Papers)}, pages 1616--1626,
  Melbourne, Australia. Association for Computational Linguistics.

\bibitem[{Szubert et~al.(2020)Szubert, Damonte, Cohen, and
  Steedman}]{szubert-etal-2020-role}
Ida Szubert, Marco Damonte, Shay~B. Cohen, and Mark Steedman. 2020.
\newblock \href {https://doi.org/10.18653/v1/2020.findings-emnlp.199} {The role
  of reentrancies in {A}bstract {M}eaning {R}epresentation parsing}.
\newblock In \emph{Findings of the Association for Computational Linguistics:
  EMNLP 2020}, pages 2198--2207, Online. Association for Computational
  Linguistics.

\bibitem[{Vaswani et~al.(2017)Vaswani, Shazeer, Parmar, Uszkoreit, Jones,
  Gomez, Kaiser, and Polosukhin}]{NIPS2017_7181}
Ashish Vaswani, Noam Shazeer, Niki Parmar, Jakob Uszkoreit, Llion Jones,
  Aidan~N Gomez, \L~ukasz Kaiser, and Illia Polosukhin. 2017.
\newblock \href
  {http://papers.nips.cc/paper/7181-attention-is-all-you-need.pdf} {Attention
  is all you need}.
\newblock In I.~Guyon, U.~V. Luxburg, S.~Bengio, H.~Wallach, R.~Fergus,
  S.~Vishwanathan, and R.~Garnett, editors, \emph{Advances in Neural
  Information Processing Systems 30}, pages 5998--6008. Curran Associates, Inc.

\bibitem[{Velickovic et~al.(2018)Velickovic, Cucurull, Casanova, Romero,
  Li{\`{o}}, and Bengio}]{velickovic2018graph}
Petar Velickovic, Guillem Cucurull, Arantxa Casanova, Adriana Romero, Pietro
  Li{\`{o}}, and Yoshua Bengio. 2018.
\newblock \href {https://openreview.net/forum?id=rJXMpikCZ} {Graph attention
  networks}.
\newblock In \emph{6th International Conference on Learning Representations,
  {ICLR} 2018, Vancouver, BC, Canada, April 30 - May 3, 2018, Conference Track
  Proceedings}.

\bibitem[{Vougiouklis et~al.(2018)Vougiouklis, Elsahar, Kaffee, Gravier,
  Laforest, Hare, and Simperl}]{VOUGIOUKLIS20181}
Pavlos Vougiouklis, Hady Elsahar, Lucie-Aimée Kaffee, Christophe Gravier,
  Frédérique Laforest, Jonathon Hare, and Elena Simperl. 2018.
\newblock \href {https://doi.org/https://doi.org/10.1016/j.websem.2018.07.002}
  {Neural wikipedian: Generating textual summaries from knowledge base
  triples}.
\newblock \emph{Journal of Web Semantics}, 52-53:1 -- 15.

\bibitem[{Wang et~al.(2019)Wang, Pruksachatkun, Nangia, Singh, Michael, Hill,
  Levy, and Bowman}]{NIPS2019_8589}
Alex Wang, Yada Pruksachatkun, Nikita Nangia, Amanpreet Singh, Julian Michael,
  Felix Hill, Omer Levy, and Samuel Bowman. 2019.
\newblock \href
  {https://proceedings.neurips.cc/paper/2019/file/4496bf24afe7fab6f046bf4923da8de6-Paper.pdf}
  {Superglue: A stickier benchmark for general-purpose language understanding
  systems}.
\newblock In H.~Wallach, H.~Larochelle, A.~Beygelzimer, F.~Alch\'{e}-Buc,
  E.~Fox, and R.~Garnett, editors, \emph{Advances in Neural Information
  Processing Systems 32}, pages 3266--3280.

\bibitem[{Wang et~al.(2018)Wang, Singh, Michael, Hill, Levy, and
  Bowman}]{wang-etal-2018-glue}
Alex Wang, Amanpreet Singh, Julian Michael, Felix Hill, Omer Levy, and Samuel
  Bowman. 2018.
\newblock \href {https://doi.org/10.18653/v1/W18-5446} {{GLUE}: A multi-task
  benchmark and analysis platform for natural language understanding}.
\newblock In \emph{Proceedings of the 2018 {EMNLP} Workshop {B}lackbox{NLP}:
  Analyzing and Interpreting Neural Networks for {NLP}}, pages 353--355,
  Brussels, Belgium. Association for Computational Linguistics.

\bibitem[{Wolf et~al.(2019)Wolf, Debut, Sanh, Chaumond, Delangue, Moi, Cistac,
  Rault, Louf, Funtowicz, and Brew}]{wolf2019huggingfaces}
Thomas Wolf, Lysandre Debut, Victor Sanh, Julien Chaumond, Clement Delangue,
  Anthony Moi, Pierric Cistac, Tim Rault, Rémi Louf, Morgan Funtowicz, and
  Jamie Brew. 2019.
\newblock \href {http://arxiv.org/abs/1910.03771} {Huggingface's transformers:
  State-of-the-art natural language processing}.

\bibitem[{Xu et~al.(2019)Xu, Hu, Leskovec, and Jegelka}]{xu2018how}
Keyulu Xu, Weihua Hu, Jure Leskovec, and Stefanie Jegelka. 2019.
\newblock \href {https://openreview.net/forum?id=ryGs6iA5Km} {How powerful are
  graph neural networks?}
\newblock In \emph{7th International Conference on Learning Representations,
  {ICLR} 2019, New Orleans, LA, USA, May 6-9, 2019}.

\bibitem[{Zhang et~al.(2019)Zhang, Sax, Zamir, Guibas, and
  Malik}]{sidetuning2019}
Jeffrey~O. Zhang, Alexander Sax, Amir Zamir, Leonidas~J. Guibas, and Jitendra
  Malik. 2019.
\newblock \href {https://arxiv.org/pdf/1912.13503.pdf} {Side-tuning: Network
  adaptation via additive side networks}.
\newblock \emph{arXiv e-prints}.

\bibitem[{Zhang et~al.(2020{\natexlab{a}})Zhang, Kishore, Wu, Weinberger, and
  Artzi}]{bert-score}
Tianyi Zhang, Varsha Kishore, Felix Wu, Kilian~Q. Weinberger, and Yoav Artzi.
  2020{\natexlab{a}}.
\newblock \href {https://openreview.net/forum?id=SkeHuCVFDr} {Bertscore:
  Evaluating text generation with {BERT}}.
\newblock In \emph{8th International Conference on Learning Representations,
  {ICLR} 2020, Addis Ababa, Ethiopia, April 26-30, 2020}.

\bibitem[{Zhang et~al.(2020{\natexlab{b}})Zhang, Guo, Teng, Lu, Cohen, Liu, and
  Bing}]{zhang-etal-2020-lightweight}
Yan Zhang, Zhijiang Guo, Zhiyang Teng, Wei Lu, Shay~B. Cohen, Zuozhu Liu, and
  Lidong Bing. 2020{\natexlab{b}}.
\newblock \href {https://doi.org/10.18653/v1/2020.emnlp-main.169} {Lightweight,
  dynamic graph convolutional networks for {AMR}-to-text generation}.
\newblock In \emph{Proceedings of the 2020 Conference on Empirical Methods in
  Natural Language Processing (EMNLP)}, pages 2162--2172, Online. Association
  for Computational Linguistics.

\end{thebibliography}
\bibliographystyle{acl_natbib}

\clearpage
\appendix

\section*{Appendices}

In this supplementary material, we detail experiments' settings and additional information about the human evaluation and graph representations.

\section{Details of Models and Hyperparameters} 
\label{appe:hyperparameters}
The experiments were executed using the version $3.3.1$ of the \emph{transformers} library released by Hugging Face \citep{wolf2019huggingfaces}. In Table \ref{tab:hyper}, we report the hyperparameters used to train the models presented in this paper. We train until the development set BLEU has not improved for 5 epochs.

\begin{table}[!h]
\centering
\resizebox{\columnwidth}{!}{
\begin{tabular}{lccc}
\toprule
         & \textbf{learning rate} & \textbf{batch size} & \textbf{beam search size} \\
\midrule

\finetune & 3e-05  & 4 & 5   \\
\finetunetop & 1e-04  & 4 & 5   \\
\finetunebottom & 1e-04  & 4 & 5   \\
\vanilladapter & 1e-04  & 4 & 5   \\
\graphadapter & 1e-04  & 4 & 5   \\
\bottomrule
\end{tabular}}
\caption{Hyperparameter settings for our methods. }
\label{tab:hyper}
\vspace{-4mm}
\end{table}

\section{Details on the Human Evaluation}
\label{appe:humaneval}
The human evaluation was conducted via Amazon Mechanical Turk. We randomly select 100 generated texts for each of the 3 sets and each adapter model (\vanilladapter, \graphadaptergcn), with a total of 600 texts to be evaluated. The annotators then rate the meaning similarity on a 1-7 Likert scale. For each text, we collect scores from 3 annotators. We use MACE~\cite{hovy-etal-2013-learning} to further improve upon these raw answers by unsupervised estimation of worker trustworthiness and subsequent recovery of the most likely score. Models are ranked according to the mean of sentence-level scores. We defined a filter for all our evaluations, allowing to participate only workers who have more than 5000 HITs approved and with an acceptance rate of 95\% or higher. The task took workers a median time of 1.6 minutes per pair of sentences. We apply a quality control step filtering workers who do not score some faked and known sentences properly or did the experiment in a very short time. 

\section{Example of Graph Linearizations}
\label{appe:exgraphline}

\begin{figure}[h]
    \centering
    \includegraphics[width=.45\textwidth]{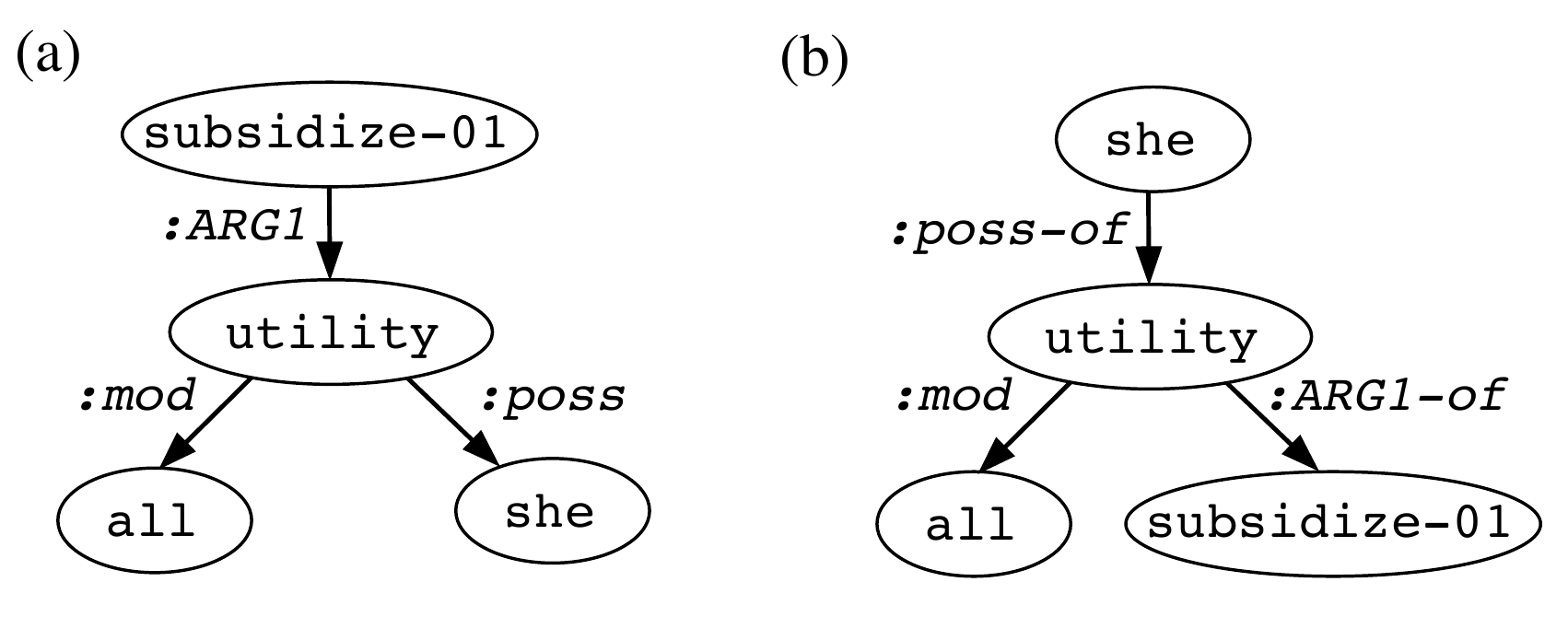}
    \caption{Two AMR graphs with the same meaning.}
    \label{fig:amrexamplesappendix}
    \vspace{-4mm}
\end{figure}

In Table~\ref{tab:sampleamrappendix}, we present three different linearizations for the same AMR graph and its corresponding reference sentence. Figure~\ref{fig:amrexamplesappendix} shows the two possible graphs that are represented by the linearizations. In particular, Figure~\ref{fig:amrexamplesappendix}a shows a graph that is represented by {\small\textsc{canon}} and {\small\textsc{reconf}} linearizations and Figure~\ref{fig:amrexamplesappendix}b shows a graph that is represented by {\small\textsc{random}}. Note that whereas the linearizations can greatly differ from each other, the graph structure for all linearizations remains very similar.

\begin{table}[!h]

	\small
	\centering
	\setlength{\tabcolsep}{3pt}
	\setlength\extrarowheight{-4pt}
	\begin{tabular}{p{7.5cm}}
		\toprule
		{\small\textsc{canon}} \\
		\\
		    (s / subsidize-01 \\
      \quad\quad :ARG1 (u / utility \\
            \quad\quad\quad\quad:poss (s2 / she) \\
            \quad\quad\quad\quad:mod (a / all))) \\
		\midrule
				{\small\textsc{reconf}} \\
		\\
		    (s / subsidize-01 \\
      \quad\quad :ARG1 (u / utility \\
            \quad\quad\quad\quad :mod (a / all) \\
            \quad\quad\quad\quad :poss (s2 / she))) \\
            
		\midrule
						{\small\textsc{random}} \\
		\\
		    (s2 / she \\
      \quad\quad :poss-of (u / utility \\
            \quad\quad\quad\quad:ARG1-of (s / subsidize-01) \\
            \quad\quad\quad\quad:mod (a / all))) \\
		\midrule
		{\small\textsc{sentence}}: Her utilities are all subsidized.\\
		\bottomrule
	\end{tabular}
	\caption{Different linearizations for an AMR graph.}
	
	\label{tab:sampleamrappendix}
	\vspace{-4mm}
\end{table}

\end{document}